\documentclass{article}

\usepackage[preprint,nonatbib]{neurips_data_2024}

\usepackage[utf8]{inputenc} 
\usepackage[T1]{fontenc}    
\usepackage{hyperref}       
\usepackage{url}            
\usepackage{booktabs}       
\usepackage{amsfonts}       
\usepackage{nicefrac}       
\usepackage{microtype}      
\usepackage[dvipsnames]{xcolor}
\usepackage{graphicx}
\usepackage{wrapfig}
\usepackage{caption}
\usepackage{amsthm}
\usepackage{amsthm}
\usepackage{newtxtext,newtxmath}
\usepackage{bm}

\usepackage{pifont}
\newcommand{\cmark}{\ding{51}}%
\newcommand{\xmark}{\ding{55}}%

\title{Open-LLM-Leaderboard: From Multi-choice to Open-style Questions for LLMs Evaluation, Benchmark, and Arena} 
        
\author{%
  Aidar Myrzakhan$^*$, Sondos Mahmoud Bsharat$^*$, and Zhiqiang Shen\thanks{Joint first author \& equal contribution.} \\
  VILA-Lab, Mohamed bin Zayed University of AI\\
  \texttt{\{aidar.myrzakhan,sondos.bsharat,zhiqiang.shen\}@mbzuai.ac.ae} \\
}

\begin{document}

\maketitle

\setcounter{footnote}{0}

\begin{abstract}
  Multiple-choice questions (MCQ) are frequently used to assess large language models (LLMs). Typically, an LLM is given a question and selects the answer deemed most probable after adjustments for factors like length. Unfortunately, LLMs may inherently favor certain answer choice IDs, such as A/B/C/D, due to inherent biases of priori unbalanced probabilities, influencing the prediction of answers based on these IDs. Previous research has introduced methods to reduce this ``selection bias'' by simply permutating options on a few test samples and applying them to new ones. Another problem of MCQ is the lottery ticket choice by ``random guessing''. The LLM does not learn particular knowledge, but the option is guessed correctly. This situation is especially serious for those small-scale LLMs\footnote{For instance, on MMLU, the random guessing accuracy is 25\%, and most small-scale LLMs obtain results around this value as shown in~\cite{shen2023slimpajama,zhang2024tinyllama}. It is difficult to distinguish which model is better under this situation.}. To address them, a more thorough approach involves shifting from MCQ to open-style questions, which can fundamentally eliminate selection bias and random guessing issues. However, transitioning causes its own set of challenges in (1) identifying suitable open-style questions and (2) validating the correctness of LLM open-style responses against human-annotated ground-truths. This work aims to tackle these significant difficulties, and establish a new LLM evaluation benchmark through entirely open-style questions. Consequently, we introduce the {\texttt{Open-LLM-Leaderboard}} to track various LLMs' performance and reflect true capability of them, such as GPT-4o/4/3.5, Claude 3, Gemini, etc. Our code and dataset are available at \url{https://github.com/VILA-Lab/Open-LLM-Leaderboard}.
\end{abstract}

\section{Introduction}

\begin{wrapfigure}{R}{7.2cm}
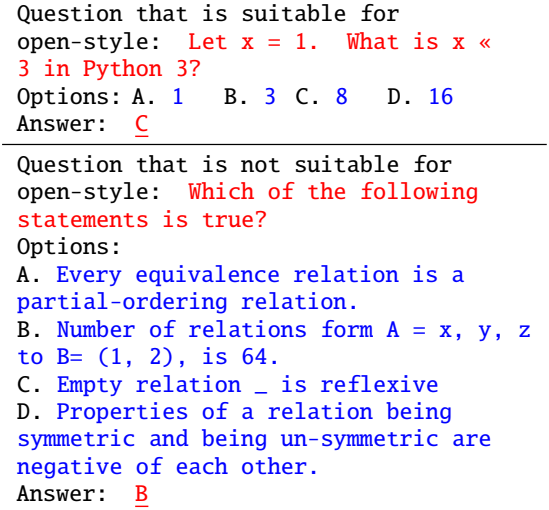

\vspace{-6mm}
\centering
    \scalebox{0.95}{
    \begin{tabular}{p{\linewidth}}
      \toprule
      \texttt{Question that is suitable for open-style: \textcolor{red}{Let x = 1. What is x << 3 in Python 3?}} \\
      \texttt{Options:} 
      \texttt{A. \textcolor{blue}{1} \ \  B. \textcolor{blue}{3}} \ \
      \texttt{C. \textcolor{blue}{8} \ \  D. \textcolor{blue}{16}} \\
      \texttt{Answer: \textcolor{red}{ \underline{C}}} \\ \midrule
      \texttt{Question that is not suitable for open-style: \textcolor{red}{Which of the following statements is true?}} \\
      \texttt{Options:} \\ \texttt{A. \textcolor{blue}{Every equivalence relation is a partial-ordering relation.}}  \\  \texttt{B. \textcolor{blue}{Number of relations form A = {x, y, z} to B= (1, 2), is 64.}}  \\
      \texttt{C. \textcolor{blue}{Empty relation \_ is reflexive}}  \\ \texttt{D. \textcolor{blue}{Properties of a relation being symmetric and being un-symmetric are negative of each other.}} \\
      \texttt{Answer: \textcolor{red}{ \underline{B}}} \\
      \bottomrule
    \end{tabular}
    }
\vspace{-2mm}
\caption{Examples of MCQ from MMLU.
}
\label{fig:mcq_example}
\vspace{-4.6mm}
\end{wrapfigure}

Large language models (LLMs) are increasingly excelling at various natural language processing tasks, including text generation~\cite{Chung_2023}, translation~\cite{pmlr-v202-zhang23m,zhu2023multilingual}, summarization~\cite{liu2023learning}, code generation~\cite{li2023starcoder,r2024code}, and chatbot interaction~\cite{openai2024gpt4}. With the rising capability, the need for a robust evaluation strategy that can accurately assess the performance of these models is becoming crucial in order to identify their true effectiveness and choose the most appropriate one for a given task. Common metrics for assessing LLMs today include relevance, frequency of hallucinations, accuracy in question answering, toxicity, and retrieval-specific metrics, among others. In the context of question-answering evaluations, prior works usually investigate the model's performance in terms of answer accuracy, courtesy, and conciseness. And multiple choice questions (MCQ) have emerged as a predominant format for such assessments, wherein a question is presented with several possible responses, and the model is required to select the most fitting choice ID, as exemplified in Figure~\ref{fig:mcq_example}. Lately, the MCQ format has seen widespread application in LLM-focused contexts, including benchmarks~\cite{hendryckstest2021,zellers-etal-2019-hellaswag,Clark2018ThinkYH} that examine LLM capabilities and automated/crowdsourcing evaluation frameworks~\cite{alpaca_eval,zheng2023judging,open-llm-leaderboard} that streamline the assessment process.

However, previous studies~\cite{zheng2024large,robinson2023leveraging} have discussed that the lack of resilience of LLMs to changes in the positioning of options stems from their tendency to exhibit biased behavior: they often favor choosing certain option IDs (such as ``Option A'') as responses, a phenomenon that is referred to as {\em selection bias}. Moreover, it shows that selection bias exists widely across various LLMs and cannot be mitigated by simple prompting skills. The underlying reason of this issue comes from the condition that the model is trained with a priori distribution that assigns more probabilistic choices to specific ID tokens. Another issue of MCQ is the ``random guessing'' that is discussed in~\cite{shen2023slimpajama}. Specifically, small models such as the 1B-level variants, may struggle to achieve reliable predictions on many benchmarks like MMLU which uses four choices as the answer candidates of the questions. Their results could resemble random choices, not truly capturing the model’s actual capabilities.

To fundamentally eliminate selection bias and random guessing in LLMs, in this work, we build an open-style question benchmark for LLM evaluation. Leveraging this benchmark, we present the {\texttt{Open-LLM-Leaderboard}}, a new automated framework designed to refine the assessment process of LLMs. This framework functions in supplement to prior evaluation frameworks such as \cite{alpaca_eval,zheng2023judging,open-llm-leaderboard} with several advantages as presented in Sec.~\ref{advantage}. However, constructing such a benchmark has two significant challenges: (1) how to determine the appropriate questions that can be effectively transformed from MCQ into open-style questions, and (2) how to establish an approach to accurately validate the correctness of the LLM's open-style answers in comparison to human-annotated ground-truths, especially in contrast to MCQ, which typically have defined single-choice standard answers.

For the first challenge of identifying the multiple-choice questions that are suitable for converting to open-style questions, we design an automatic coarse-to-fine selecting protocol through customized prompts and multi-stage filtering process. Specifically, in the first stage, we use the binary classification to filter the questions with high confidence as the positive pool and others are assigned as negative. Our second stage uses a soft scoring method (1-10 ratings) to judge the suitability of the questions for the open-style from the questions that are clarified as negative in the first stage. For the second challenge of evaluating the correctness of the LLM's open-style answers in comparison to human-annotated ground-truths, we further design a task-specific prompt and leverage GPT-4 to examine if the response is correct. To validate the accuracy of the automatic evaluation strategy, we randomly sample 100 results and manually check the automatic evaluation results with the corresponding responses, and confirm that it is reliable with an error rate of less than 5\%. 

In our end-to-end assessment of the LLM evaluation and ranking process, we conduct a comprehensive analysis on the well-recognized LLMs, including GPT-4o, GPT-4, ChatGPT, Claude-3 Opus, Gemini-Pro and Mistral-Large. Our benchmarking results indicate that GPT-4o currently holds the position as the strongest LLM. We further provide a small regime LLM leaderboard targeting at LLMs smaller than 3B. Moreover, our study demonstrates a high correlation between the rankings produced by our open-style benchmark and those derived from user-based evaluations or direct human assessments.

\section{Related Work}
\textbf{Large Language Models (LLMs).} Recent advancements in LLMs, such as GPT-3 \cite{NEURIPS2020_1457c0d6} and GPT-4~\cite{openai2024gpt4} have had a significant impact in the field of natural language processing and have found widespread application across various domains. It has indeed initiated a kind of chain reaction within the community and beyond. As each new iteration of LLMs demonstrates enhanced capabilities, organizations and researchers across various sectors are motivated to develop their own models, such as LLaMA \cite{touvron2023llama, touvron2023llama2}, Gemini \cite{geminiteam2023gemini}, and Claude \cite{claude}, or find innovative ways to improve existing LLMs through instruction tuning, like Alpaca \cite{alpaca}, and Vicuna \cite{vicuna2023}. 

\textbf{Multiple Choice Questions (MCQ).} In the realm of LLM research, MCQ has become a pivotal tool for evaluating and enhancing the capabilities of these models. Notable datasets like the MMLU \cite{hendryckstest2021}, HellaSwag \cite{zellers-etal-2019-hellaswag}, and ARC \cite{Clark2018ThinkYH} have been instrumental in this regard. Their diverse assessment of broad knowledge and commonsense reasoning help in benchmarking the depth and versatility of LLMs in understanding, reasoning, and applying knowledge across various domains. MCSB~\cite{Robinson2022LeveragingLL} introduces a natural prompting strategy for LLMs, which presents questions and answer choices together, allowing the model to explicitly compare options. 

\textbf{Bias in LLMs.} Selection bias, a specific form of bias relevant to the evaluation of LLMs through MCQ, has garnered attention due to its understated and widespread impact. A series of works \cite{zheng2023judging, pezeshkpour2023large, zheng2024large, wang2023large} have shown that LLMs may develop a propensity to favor certain answer choices based on their position or encoding, such as the alphabetical ordering of A/B/C/D in MCQ. This phenomenon can lead to skewed evaluation results, misrepresenting a model's true understanding and reasoning capabilities.

\section{Approach}
\begin{figure}[t]
  \centering
    \includegraphics[width=0.95\linewidth]{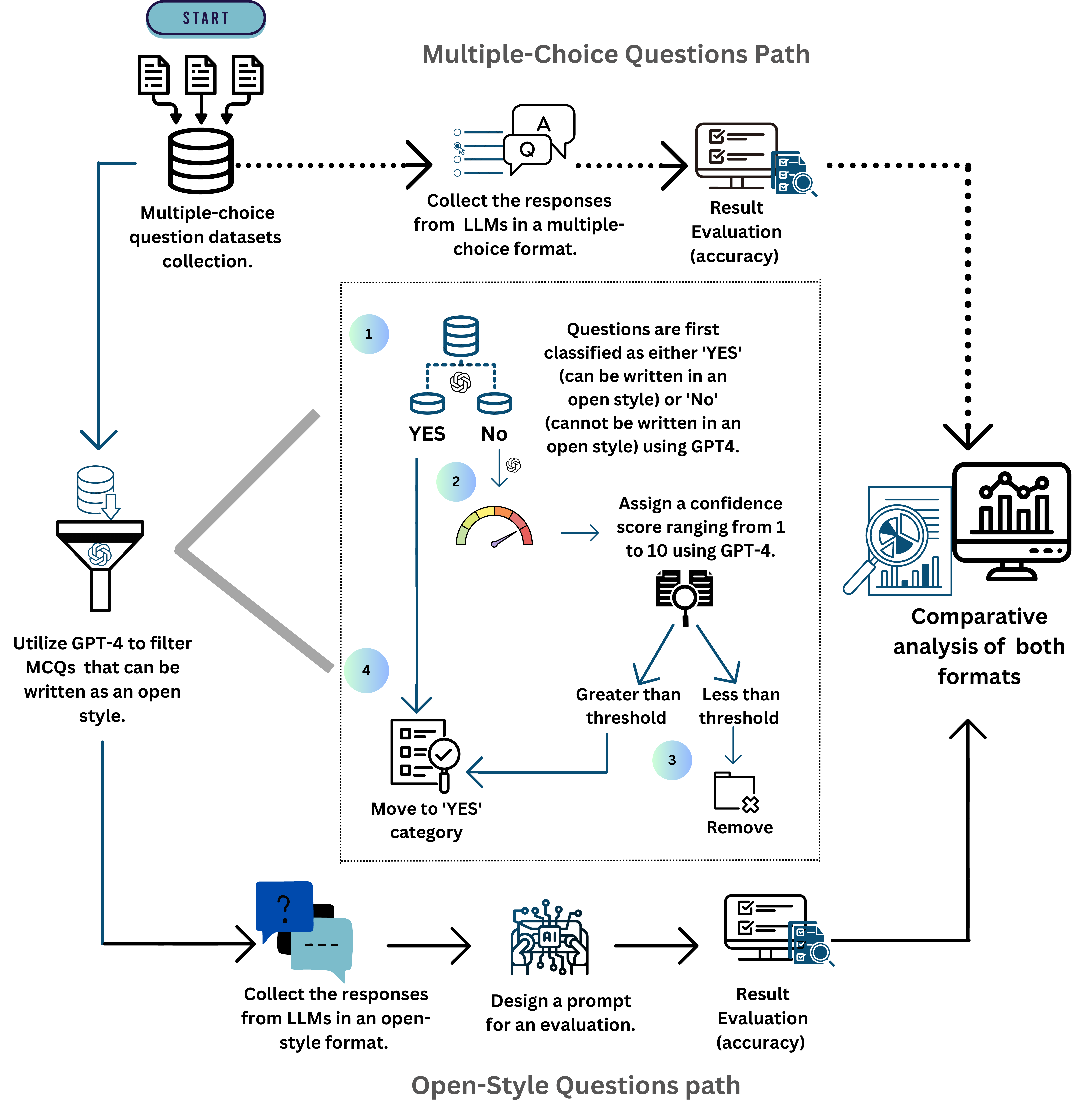}
    \vspace{-0.1in}
  \caption{An overview of a dual-path evaluation pipeline for LLMs, starting with the collection of MCQ datasets. It branches into two paths, with the MCQ path proceeding directly from response collection to evaluation, while the open-style path passes through an additional filtering phase. After evaluation, both paths converge in a comparative analysis.} 
  \label{pipeline}
  \vspace{-0.2in}
\end{figure}

\subsection{Defining Open-style Questions}

Open-style questions, aka open-ended questions, require the model to generate an answer without being constrained by a set of predetermined choices. In the context of LLM evaluation, these questions are designed to assess the model's ability to generate coherent, relevant, and contextually appropriate responses based on the input query. While multiple-choice questions can efficiently assess specific factual knowledge and comprehension, open-style questions offer a deeper insight into the LLM’s generative capabilities, understanding of context, and ability to engage with complex tasks. Also, open-style questions can avoid the inherent selection bias and random guessing weaknesses compared to multiple-choice questions.

\subsection{Automatic Open-style Question Filtering and Generation}

\noindent{\bf Multi-stage Filtering and Postprocessing via Coarse-to-fine Process.}  Our proposed multi-stage filtering approach consists of four main steps to streamline the conversion: (1) Initially classify datasets as either convertible or non-convertible. (2) Assign each question a confidence score to indicate the likelihood that it can be framed as an open-style question. (3) Exclude questions with confidence scores below a specified threshold and classified as non-convertible. (4) Combine questions that are labeled as non-convertible but have high confidence scores with those labeled as convertible.

\noindent{\bf Stage1: Preliminary Filter using Binary Classification}. Considering that the structure of MCQ varies, converting them into an open-style format is not always possible, particularly because certain questions are strongly linked to their choices. For instance, questions formulated as ``Which one of the following is true'' or ``All except'' or ``Which of these''. Such questions are typically unsuitable for conversion into an open-style format since the absence of the options could change the question’s core, resulting in incomplete questions.

\begin{table}[h]
\vspace{-0.1in}
\caption{Prompt design for two-stage filtering and post verification.}
\centering
\resizebox{1.0\textwidth}{!}{
\begin{tabular}{c}
\bf Stage One: Coarse Filtering Prompt \\
\begin{tabular}[c]{@{}p{18cm}@{}}
{\texttt{"""Your task is to review a series of multiple-choice questions and evaluate their
ability to be answered without the provided answer choices. For questions that
begin with an incomplete sentence (e.g., "During swallowing, ..."), use your
knowledge to attempt to complete the sentence accurately. For direct questions
that ask for specific information or identification (e.g., "Which of the following
structures is part of the small intestine?"), assess whether the question is
formulated clearly enough that an informed answer can be given without seeing
the multiple-choice options. For mathematical or analytical questions (e.g., "Find
all cosets of the subgroup 4Z of 2Z"), determine if the question provides enough
context and information for a solution to be formulated without additional
options. }

\vspace{\baselineskip}
\texttt{\bf Please follow this format for your evaluation:} 
\vspace{\baselineskip}

\texttt{QUESTION: [Insert the question here]
VERDICT: Respond with "YES" if the question is clear and can be directly answered
based on its content alone, or "NO" if it relies on the answer choices to be
understood or answered. Your response should include only the verdict without
any justification or reasoning."""}} \end{tabular} \\ \midrule
  \bf  Stage Two: Fine-grained Filtering Prompt \\
\begin{tabular}[c]{@{}p{18cm}@{}}
\texttt{
You will assign a numerical score from 1 to 10 based on how confidently it can
be answered without the choices. The scoring criteria are as follows:}
\vspace{\baselineskip}

\texttt{
1: The question is entirely dependent on its choices for an answer, making it
impossible to answer without them. Example: `Which of the following statements
is correct?'}
\vspace{\baselineskip}

\texttt{
10: The question can be easily and confidently answered based solely on the
question stem, without any need to refer to the provided options. Example: `What
is the first law of thermodynamics in physics?'
Intermediate Scores: }
\vspace{\baselineskip}

\texttt{
2-4: The question stem gives very little information and is highly reliant on the
choices for context. Example: `Which of these is a prime number?'
5: The question provides some context or information, that gives a moderate
possibility to answer the question. Example: `Which of the following best describes
the structure that collects urine in the body?'}
\vspace{\baselineskip}

\texttt{
6: The question provides a good amount of context or information, that gives a
moderate possibility to answer the question. Example: `Statement 1 | A factor
group of a non-Abelian group is non-Abelian. Statement 2 | If K is a normal
subgroup of H and H is a normal subgroup of G, then K is a normal subgroup of G.'}
\vspace{\baselineskip}

\texttt{
7: The question provides a good amount of context or information, that gives a high
possibility to answer the question. Example:  `The element (4, 2) of Z\_12 x Z\_8 has
order'}
\vspace{\baselineskip}

\texttt{
8-9: The question provides a good amount of context or information, that gives a
high possibility to answer the question. Example: `A "dished face" profile is often
associated with'}
\vspace{\baselineskip}

\texttt{
ONLY GIVE THE VALUE BETWEEN 1-10 AS YOUR ANSWER. DO NOT INCLUDE ANY
OTHER INFORMATION IN YOUR RESPONSE}
\vspace{\baselineskip}

\texttt{
Example Format:}

\texttt{
QUESTION: question here}

\texttt{
VERDICT: value in [1-10] here}
\end{tabular} \\ \midrule

   \bf  GPT-4 Prompt for Verification \\
\begin{tabular}[c]{@{}p{18cm}@{}}
\texttt{
"""Evaluate the answer of a AI model to a question.
You will be provided with the question, the AI model's answer, and the correct answer. Your task is to evaluate the AI model's response and determine whether it is Correct or Incorrect.}
\vspace{\baselineskip}





\vspace{\baselineskip}

\texttt{
Grade the AI model answers based ONLY on their factual accuracy. It is OK if the AI model answer contains more information than the true answer, as long as it does not contain any conflicting statements. Otherwise, it should be marked as Incorrect. Ignore differences in punctuation and phrasing between the AI model's answer and the true
answer.}
\vspace{\baselineskip}

\texttt{
Example Format:}

\texttt{
QUESTION: question here}

\texttt{
STUDENT ANSWER: student's answer here}

\texttt{
TRUE ANSWER: true answer here}

\texttt{
GRADE: Correct or Incorrect here
}

\texttt{
\noindent{Your response should include only the verdict without any justification or
reasoning."""}}
\end{tabular}
\end{tabular}
}
\vspace{-0.1in}
\label{tab:my_label}
\vspace{-0.15in}
\end{table}

 To effectively handle this challenge of identifying whether multiple-choice questions are suitable for open-style conversion, we leverage the power of prompting techniques to create a customized classification prompt as shown in Table \ref{tab:my_label}. In the prompt, we integrate different types of questions from different datasets to demonstrate how an LLM may evaluate each question to be written in an open-style way, eventually classifying them as convertible ``YES'' or non-convertible ``NO''. It will determine whether a question provides a clear context and information without relying on the provided options or not. In the prompt we integrate different types of questions from different datasets to demonstrate how an LLM like GPT-4 may evaluate each question to be written in an open-style way, eventually classifying them as convertible ``YES'' or non-convertible ``NO''. We set the prompt to eliminate any additional explanations, by stating that ``Your response should include only the verdict without any justification or reasoning.'' This guarantees that the answer to each inquiry is conveyed concisely as ``YES'' or ``NO''.
 
To understand our initial filtering results, we conduct an error analysis manually by selecting 100 questions in the ``YES'' and ``NO'' pools separately. In the samples classified as ``YES'', we find that only around 5\% of the questions are false positive cases, verifying a low misclassification error for the positive question selection by our filtering strategy. Conversely, within the ``NO'' sample, around 40\% of the questions are actually suitable for open-style questions but mistakenly classified as negative. This situation often arises from questions that include phrases like ``Which of''. Similarly, questions involving true/false statements, sentence completions, or fill-in-the-blanks are also sometimes inappropriately classified as non-convertible. This analysis motivates us to develop a cascaded fine-grained stage to further filter more positive questions in ``NO'' pool using particular prompts, as described in the following Stage 2 process.

 \noindent{\bf Stage2: Confidence Score Assignment}. As we aim to overcome the issue of classifying questions with specific patterns as non-convertible, we introduce a second stage of filtering centered on confidence score assignment. This involves instructing the large language model to assign a confidence score on a scale from 1 to 10, reflecting the possibility of the question being written in an open-style format. Since a significant number of questions are unsuitable for an open-style format, categorized as ``NO'' and have a confidence score below 5, \noindent{\bf we set a confidence score threshold to be 5}. Therefore, questions classified as non-convertible with a confidence score lower than this threshold are excluded, while those remaining above the threshold and those initially classified as convertible are   \noindent{\bf moved into the ``YES'' category to be converted to an open-style format}.
 
\subsection{Open-style Question Answer Evaluation}
\vspace{-0.05in}
After establishing a set of convertible questions from various datasets and obtaining their responses from several LLMs, there arises a need to evaluate these questions. Given that our ground truth answers are based on the MCQ format with defined answers, it necessitates a method for efficiently and accurately validating the correctness of responses to open-style questions. To this end, we design a customized prompt, as shown in Figure \ref{pipeline} that utilizes the correct MCQ answer as the ground truth to determine if the open-style responses are correct or incorrect by the prediction \(\hat y\):
\begin{equation}
    \hat y = \texttt{LLM}_\texttt{e}(\text{prompt}(q, \hat{a}, a))
\end{equation}
where $\hat y$ represents the prediction and $\texttt{LLM}_\texttt{e}$ is the LLM evaluator. \(q\), \(\hat{a}\) and  \(a\) represent the question, LLM generated answer, and correct answer from MCQ, respectively, and the prompt is provided in Table~\ref{tab:my_label} of Appendix.
While these open-style answers are evaluated based on the MCQ's ground truth, issues of misevaluation might arise. This includes scenarios where a response is inaccurately classified as correct simply because it contains certain keywords also found in the ground truth. To tackle this issue we include specific phrases in the prompt. These phrases, such as ``as long as it does not contain any conflicting statements'', ensure that a response is not automatically classified as correct based on the presence of a keyword, avoiding incorrect markings when the response contradicts the correct answer. Additionally, to prevent the exclusion of correct answers that incorporate extra information, we incorporate the phrase ``It is OK if the AI model's answer contains more information than the true answer''. Furthermore, we highlight that minor differences in punctuation and phrasing between the open-style responses and the ground truth answers should not lead to their being classified as incorrect. To see the correctness of the LLM judgement we take the randomly drawn 100 responses from all models. The human evaluation process for our study was conducted by the authors themselves. The agreement between the LLM evaluations and those of a human evaluator was quantitatively assessed using Cohen's kappa \cite{doi:10.1177/001316446002000104}, which yielded a score of 0.83. This substantial kappa score\footnote{
The Kappa score is a statistical measure of inter-rater agreement for categorical items, defined by the equation: 
\(\kappa = \frac{P_o - P_e}{1 - P_e}\)
where \( P_o \) is the observed agreement and \( P_e \) is the expected agreement by chance.} verifies that the LLM's ability to determine the correctness of responses aligns closely with human judgment, demonstrating strong reliability in its evaluation process.

\section{An Open-style Question Benchmark (OSQ-bench)}

\subsection{Statistics and Distributions} \label{stats}

Table \ref{tab:stat_os} describes the basic statistics of the dataset questions that are suitable for answering in open-style format. In total, we have evaluated 42K questions from 9 different datasets and more than 23K of them are classified as appropriate for open-style answering. 

\begin{table}[ht]
\centering
\vspace{-0.15in}
\caption{Statistics on open-style questions across different datasets.}
\resizebox{0.8\textwidth}{!}{
\begin{tabular}{@{}lcccc@{}}
\toprule
\textbf{Benchmarks} & \textbf{\#Evaluated} & \textbf{\#Open-Style} & \textbf{Average Question Length (words)} \\ \midrule
MMLU & 14,042 & 7,784 &  36.6 \\
ARC & 3,428 & 3,118 & 21.1 \\ 
MedMCQA  &  4,183    & 2,318   & 14.1 \\
CommonsenseQA &  1,221   & 710      &  13.1 \\
Race & 4,934 & 3,520 & 10.0 \\
OpenbookQA & 1,000 & 491 & 10.3 \\
WinoGrande & 1,267 & 1,267 & 19.1 \\
HellaSwag & 10,042 & 3,915 & 40.1\\
PIQA & 1,838 & 696 & 7.1\\
\textbf{Overall} &  \textbf{41,955} & \textbf{23,839} & \textbf{19.05} \\
\bottomrule
\end{tabular}
}
\label{tab:stat_os}
\vspace{-0.1in}
\end{table}

\subsection{Diversity}
\begin{wrapfigure}{l}{0.48\textwidth}
\vspace{-0.25in}
\centering
\includegraphics[width=0.5\textwidth]{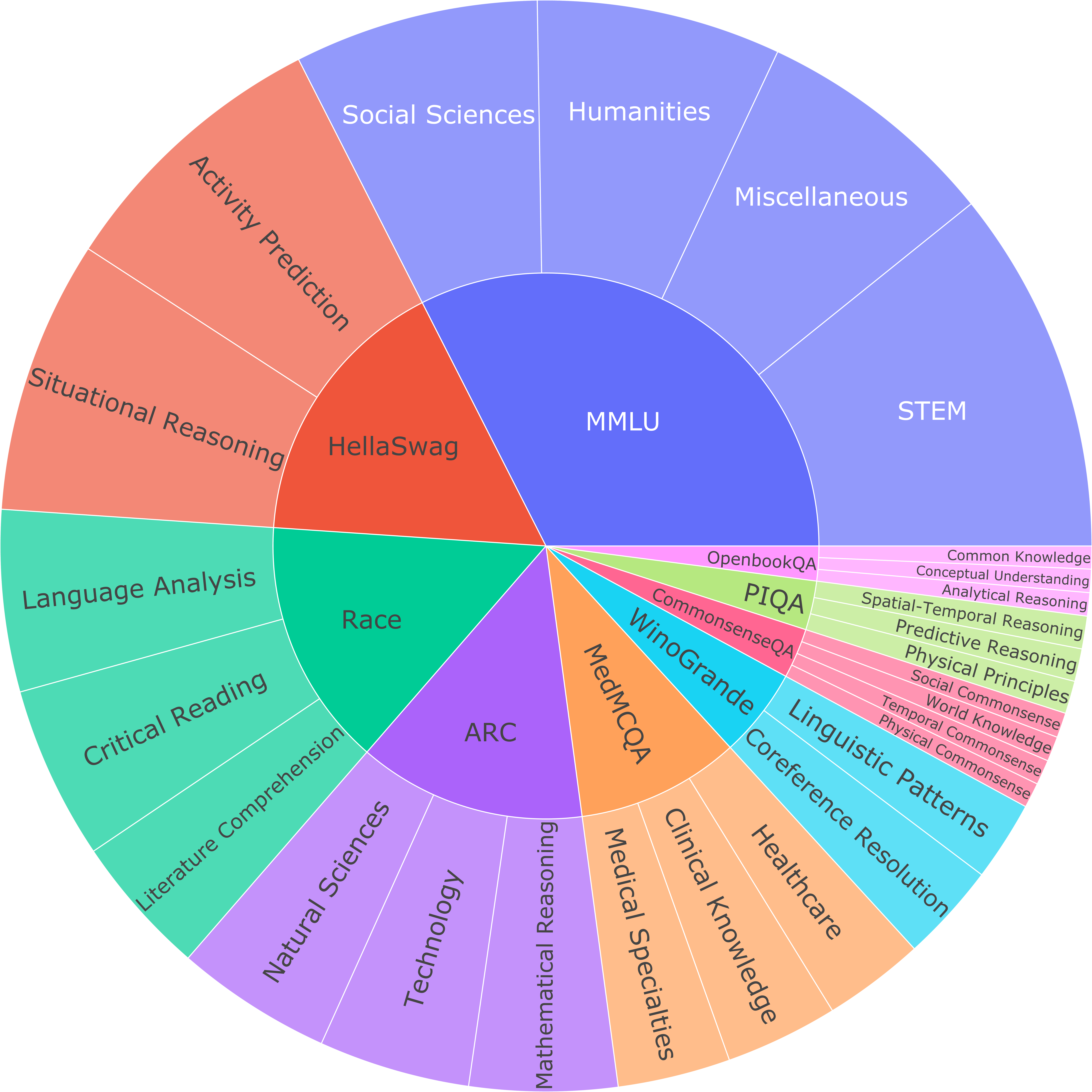}
\vspace{-0.2in}
\caption{Diversity and distribution of used datasets for our {\bf OSQ-bench}.}
\label{fig:data_distribution}
\vspace{-0.45in}
\end{wrapfigure}

\vspace{-0.05in}
Our investigation into the diversity of questions within our benchmark is foundational for understanding the landscape of open-ended question answering. To comprehensively assess the breadth of question diversity, we have conducted a systematic categorization of the question types sourced from an array of distinct datasets. From the total initial pool of 41,955 questions, we refine the selection to 23,839 questions, ensuring that each one is conducive to open-ended responses. The distribution of those questions is illustrated in Figure \ref{fig:data_distribution}, which segments the data into several domains based on the content of the questions. The segmentation of the plot underscores the interdisciplinary nature of our dataset. It features a broad spectrum of categories such as literature and reading comprehension, commonsense reasoning, domain-specific (medicine, STEM, and etc), and multi-topic knowledge. Also, Table \ref{tab:stat_os} demonstrates the diversity of question length used for the benchmark.

\vspace{-0.05in}
\subsection{Quality}
Our newly developed benchmark, curated from widely recognized datasets, stands out by focusing on questions suitable to open-style answering, i.e., a format that demands a deep understanding and an ability to generate informative, unrestricted responses. Given that the datasets from which these questions originate are widely utilized and highly recognizable within the research community, it follows that the questions have good quality to assess the models' capabilities. Moreover, due to the thorough filtering process it has undergone, it results in a low false positive rate (questions not suitable for open-style that are classified as suitable) of around 5\%. This indicates that the vast majority of questions categorized as suitable for open-style answers indeed meet the criteria. 

\vspace{-0.15in}
\subsection{Property and Advantage} \label{advantage}
\vspace{-0.05in}
As shown in Table~\ref{tab:comparison}, our leaderboard exhibits several advantages: first is the debiased results compared to the MCQ-based leaderboard, which has been discussed thoroughly. Another advantage is the faster and cheaper evaluation over crowduser-based leaderboards. Our results and rankings can be generated automatically without any human intervention.

\begin{table}[h]
\begin{center}
\vspace{-0.15in}
\caption{Comparison with different LLM leaderboards. ``Biased'' indicates the selection bias.}
\label{tab:comparison}
\resizebox{1.0\textwidth}{!}{
\begin{tabular}{lcccc}
\toprule
\multicolumn{1}{c}{\bf Leaderboard}  &\multicolumn{1}{c}{\bf Types} & \bf Diversity & \bf  Biased & \bf Evaluation \\
\midrule
Huggingface Leaderboard~\cite{open-llm-leaderboard}     & Multiple Choices Questions & High & \textcolor{red}{\cmark} & Automatically \\
AlpacaEval Leaderboard~\cite{alpaca_eval}             & Human Questions\&Feedback & Low  & \textcolor{green}{\xmark} & GPT-4 \\
Chatbot Arena Leaderboard~\cite{zheng2023judging}     &  Human Questions\&Feedback & Low & \textcolor{green}{\xmark} & GPT-4/Crowdusers \\
\bf Open-LLM-Leaderboard (Ours)             & Open Style Questions  & High & \textcolor{green}{\xmark} & GPT-4 \\
\bottomrule
\end{tabular}
}
\end{center}
\vspace{-0.15in}
\end{table}

\section{Experiments} \label{experiments} \label{exp}

\begin{figure}[t]
    \centering
    \includegraphics[width=0.99\linewidth]{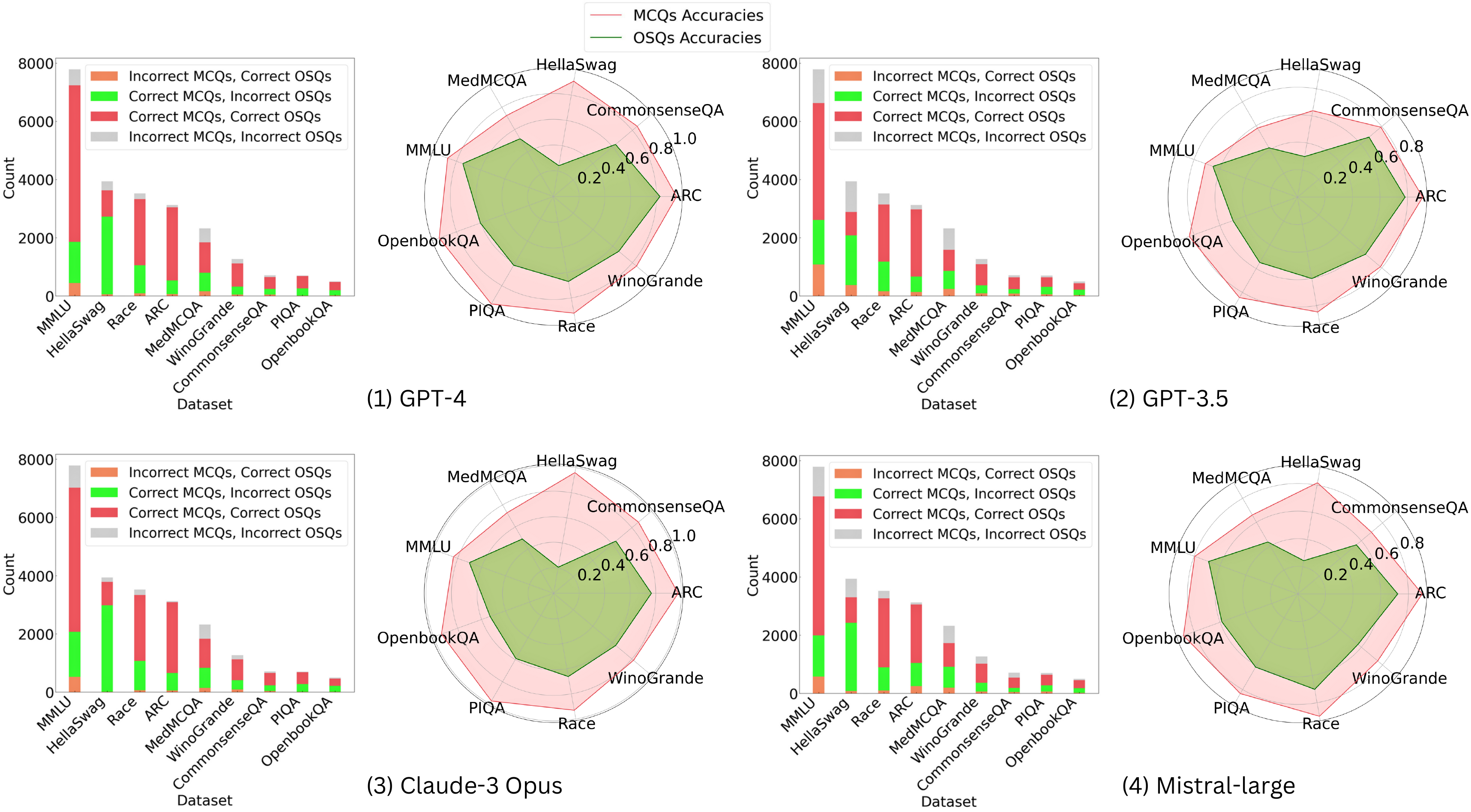}
    \vspace{-0.05in}
    \caption{Performance comparison of various LLMs on multiple-choice (MCQ) and open-style questions (OSQ) across different datasets. The bar graphs on the left show the counts of correct and incorrect responses (\textcolor{Orange}{\xmark} MCQ vs. \textcolor{Orange}{\cmark} OSQ; \textcolor{green}{\cmark} MCQ vs. \textcolor{green}{\xmark} OSQ; \textcolor{WildStrawberry}{\cmark} MCQ vs. \textcolor{WildStrawberry}{\cmark} OSQ; \textcolor{lightgray}{\xmark} MCQ vs. \textcolor{lightgray}{\xmark} OSQ), while the radar charts on the right illustrate the accuracy comparisons between MCQ and OSQ for each language model (\textcolor{pink}{Pink} is the MCQ accuracy and \textcolor{LimeGreen}{LimeGreen} is the OSQ accuracy).} 
  \label{TF1}
  \vspace{-0.13in}
\end{figure}

\begin{figure}[t]
    \centering
    \includegraphics[width=0.75\linewidth]{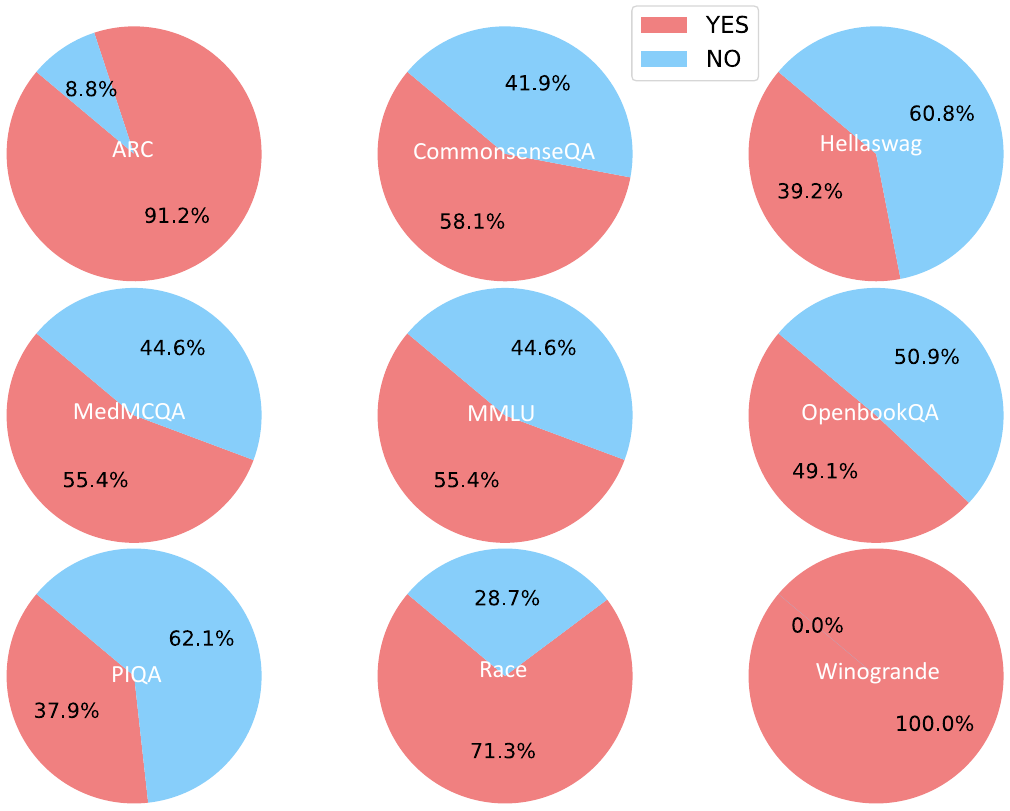}
  \vspace{-0.05in}
  \caption{Percentage of convertible MCQ to open style questions on various datasets.}
  \label{TF}
  \vspace{-0.15in}
\end{figure}

\vspace{-0.05in}
\subsection{Models} 
\vspace{-0.05in} 
We generate responses from LLMs of different sizes. The large-scale LLMs: \texttt{gpt-3.5-turbo}, \texttt{gpt-4-1106-preview}, \texttt{gpt-4o} \cite{openai_chat},  \texttt{claude-3-opus-20240229} \cite{claude_chat}, \texttt{mistral-large-latest} \cite{mistral_chat}, \texttt{gemini-pro} \cite{gemini_pro}, and \texttt{llama3} \cite{llama3modelcard}. We use the commercial APIs to collect responses from all of these models. The small-scale LLMs: \texttt{qwen1.5} \cite{qwen}, \texttt{gemma} \cite{gemmateam2024gemma}, \texttt{SlimPajama-DC} \cite{shen2023slimpajama}, \texttt{RedPajama} \cite{mosaicml_mpt1b_redpajama_200b}, \texttt{OLMo} \cite{Groeneveld2023OLMo}, \texttt{Pythia} \cite{biderman2023pythia}, \texttt{TinyLlama} \cite{zhang2024tinyllama}, \texttt{OPT} \cite{zhang2022opt}, \texttt{GPT-Neo} \cite{gpt-neo}, and \texttt{Cerebras-GPT} \cite{dey2023cerebrasgpt}. All of the small-scale model responses are collected using Huggingface \cite{DBLP:journals/corr/abs-1910-03771} and lm-evaluation-harness framework \cite{eval-harness} with 4$\times$ 4090 RTX GPUs.

\subsection{Datasets}\label{datas}
\vspace{-0.05in}
We present a brief overview of used datasets, highlighting their distinctive characteristics and the specific aspects they aim to evaluate. MMLU \cite{hendryckstest2021}, ARC \cite{Clark2018ThinkYH}, and MedMCQA \cite{pmlr-v174-pal22a} stand out with their comprehensive range of tasks spanning across various disciplines. PIQA \cite{Bisk_Zellers_Le}, CommonsenseQA \cite{talmor-etal-2019-commonsenseqa}, OpenBookQA \cite{mihaylov-etal-2018-suit}, and HellaSwag \cite{zellers-etal-2019-hellaswag} focus on the different aspects of commonsense reasoning, such as physical interaction, everyday concepts, and their interrelations.  RACE \cite{lai-etal-2017-race} provides a source of reading comprehension challenges. WinoGrande \cite{10.1145/3474381} is designed to test the model on resolving coreferences and understanding nuanced relationships in text. This dataset with its unique fill-in-a-blank tasks, inherently aligns with open-ended question formats, negating the need for our multi-stage filtering process. For other datasets, questions are filtered using \texttt{gpt-4-0125-preview} using prompts from Table \ref{tab:my_label}. The prompts for both MCQ and OSQ on each dataset are in Appendix \ref{pfdd}.

\subsection{Evaluation} \label{eval}
\vspace{-0.05in}
\begin{table}[h]
\centering
\caption{Comparison of multiple choice (MCQ) and open style questions (OSQ) accuracy.}
\resizebox{1\textwidth}{!}{
\begin{tabular}{lcc|cc|cc|cc|cc}
\toprule
Dataset   & \multicolumn{2}{c}{\textbf{GPT-4}}  & \multicolumn{2}{c}{\textbf{GPT-3.5}} & \multicolumn{2}{c}{\textbf{Gemini Pro}} & \multicolumn{2}{c}{\textbf{Claude-3 O}} & \multicolumn{2}{c}{\textbf{Mistral L}} \\

\cmidrule(lr){2-3} \cmidrule(lr){4-5} \cmidrule(lr){6-7} \cmidrule(lr){8-9} \cmidrule(lr){10-11}
               & MCQ          & \multicolumn{1}{r}{OSQ}             & MCQ          & \multicolumn{1}{r}{OSQ}            & MCQ       & \multicolumn{1}{r}{OSQ} & MCQ       & \multicolumn{1}{r}{OSQ} & MCQ & \multicolumn{1}{r}{OSQ} \\
\midrule
MMLU            & 87.28        & 74.77   &  71.25       & 65.38     & 65.71     & \multicolumn{1}{r|}{56.04}   & 83.52     & \multicolumn{1}{r|}{70.23} & 79.50     & \multicolumn{1}{r}{68.76}\\
ARC             & 95.54       & 82.68   & 90.64        &  78.42   & 90.96     & \multicolumn{1}{r|}{72.35}   & 97.50     & \multicolumn{1}{r|}{75.47} & 89.96     & \multicolumn{1}{r}{72.32}\\
HellaSwag       & 90.98        & 24.35   & 63.84        &  29.99   & 69.05     & \multicolumn{1}{r|}{25.69}   & 96.04     & \multicolumn{1}{r|}{20.79} & 81.78     & \multicolumn{1}{r}{24.47}\\
WinoGrande      & 84.14        & 66.22   & 78.77        & 64.56    & 66.85     & \multicolumn{1}{r|}{56.35}   & 81.69     & \multicolumn{1}{r|}{63.54} & 75.45     & \multicolumn{1}{r}{56.83}\\
PIQA            & 96.41        & 61.64   & 84.34       &  54.89   & 83.33     & \multicolumn{1}{r|}{47.70}   & 97.41     & \multicolumn{1}{r|}{59.05} & 83.33     & \multicolumn{1}{r}{61.21}\\
CommonsenseQA   & 84.93        & 62.96   & 79.15        &  67.89   & 66.62     & \multicolumn{1}{r|}{50.56}   & 86.76     & \multicolumn{1}{r|}{63.66} & 69.58     & \multicolumn{1}{r}{55.35}\\
Race            & 92.02        & 67.05   & 84.80        & 60.11    & 87.73     & \multicolumn{1}{r|}{61.02}   & 93.04     & \multicolumn{1}{r|}{66.22} & 89.97     & \multicolumn{1}{r}{70.17}\\
MedMCQA         & 72.65        & 51.81   & 58.02        & 41.42  & 58.02     & \multicolumn{1}{r|}{35.89}   & 72.91 & \multicolumn{1}{r|}{49.14} & 66.05     & \multicolumn{1}{r}{43.44}\\
OpenbookQA      &  94.30       & 60.29   & 83.71        &  49.90   & 86.97     & \multicolumn{1}{r|}{52.55}   & 93.48     & \multicolumn{1}{r|}{52.95} & 88.19     & \multicolumn{1}{r}{58.66}\\
\textbf{Average} & 88.69 & 61.31 & 78.28 & 56.95 & 75.03 & 50.91 & 90.26 & 57.89 & 80.42 & 56.80 \\
\bottomrule
\end{tabular}
}
\label{compare-mcq-os}
\vspace{-0.1in}
\end{table}

Our assessment approach for both MCQ and OSQ aligns with widely recognized evaluation frameworks and leaderboards for LLMs. The evaluation of MCQ is conducted utilizing the OpenAI Evals framework \cite{openai_evals} with the zero-shot setting, which involves comparing the generated response with the ground truth ID. In contrast, for evaluating responses to open-ended questions, we employ the \texttt{gpt-4-0125-preview} model to determine the correctness of responses generated by LLMs relative to a pre-established ground truth answer from the dataset using the prompt from Table \ref{tab:my_label}. 

The results in Table \ref{compare-mcq-os} and Figure \ref{TF1} are based on filtered questions. They show that every model experiences a significant drop in the accuracy for OSQ compared to MCQ. On average, the accuracy of OSQ is lower than MCQ by about 25\% for all models.  This result can correlate with our concern that the model will ``randomly guess'' to correct choices but it cannot answer. This discrepancy in performance between OSQ and MCQ is not necessarily a negative reflection of the models' overall capabilities. Instead, it can be viewed as a true comparison of the models' abilities to process and understand diverse types of questions.

The most significant difference in models between OSQ and MCQ is observed for Claude-3 Opus, by 31\%. The dataset with the largest fall between MCQ and OSQ is HellaSwag. This is because of the type of questions in this dataset. It asks to choose the most plausible continuation for the scenarios presented. Evaluating the OSQ responses of LLMs against the ground truth in this dataset presents a significant challenge due to the different plausible completions. It means that a multitude of valid and contextually appropriate answers can exist, which makes it difficult to evaluate with single-choice ground truth. This contrasts with WinoGrande, which consists of questions that require fill-in-the-blank in sentences with correct words. As a result, HellaSwag does not seem well-suited for open-style questions, and we have chosen to omit it from our final leaderboard.

\begin{table}[t]
\centering
\caption{Open-LLM Leaderboard for Large-scale Models. WG, CSQA, OBQA, and HS represent WinoGrande, CommonsenseQA, OpenbookQA, and HellaSwag respectively. We did not include HellaSwag results in the overall accuracy as the evaluation difficulties mentioned in Sec.~\ref{eval}. }
\resizebox{1\textwidth}{!}{
\begin{tabular}{l|c|ccccccccc}
\toprule
   & \textbf{Overall} & \textbf{MMLU}  & \textbf{ARC} & \textbf{WG} & \textbf{PIQA} & \textbf{CSQA} & \textbf{Race} & \textbf{MedMCQA} & \textbf{OBQA}  & \textcolor{gray}{\textbf{HS}} \\
\midrule
\textbf{GPT-4o} \raisebox{-0.04in}{\includegraphics[width=0.03\linewidth]{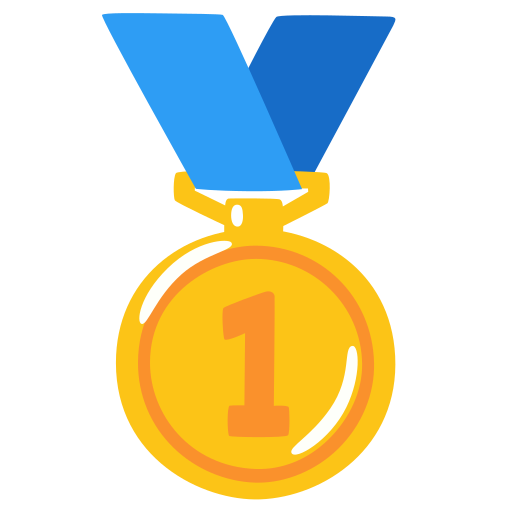}} & \textbf{70.15} & \textbf{79.09} & \textbf{86.31}& \textbf{72.22} & 60.34 & \textbf{70.28} & 67.87 & \textbf{57.85} & \textbf{67.21}  & \textcolor{gray}{--} \\
\textbf{GPT-4-1106-preview} \raisebox{-0.04in}{\includegraphics[width=0.03\linewidth]{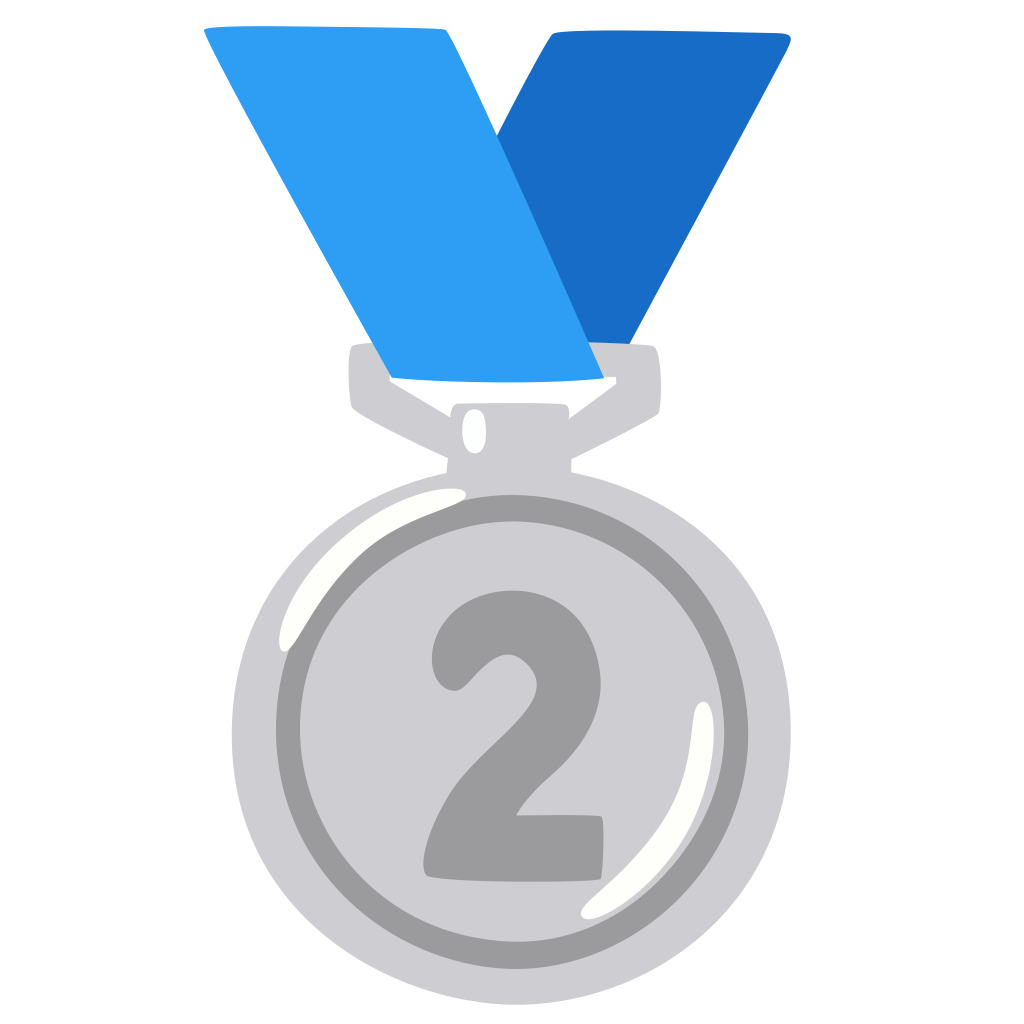}} & 65.93 & 74.77 & 82.68 & 66.22 & \textbf{61.64} & 62.96 & 67.05 & 51.81 & 60.29  & \textcolor{gray}{24.35} \\
\textbf{Claude-3 Opus}\raisebox{-0.04in}{\includegraphics[width=0.03\linewidth]{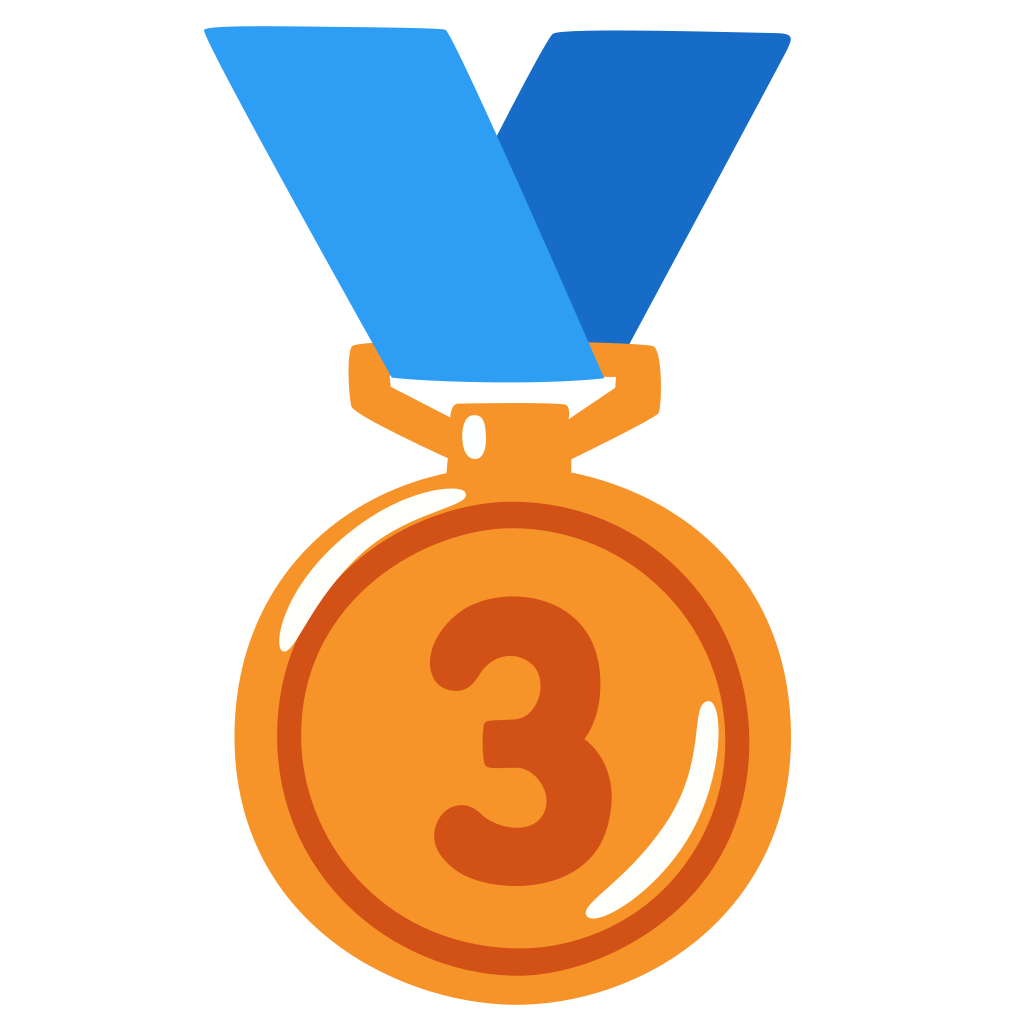}} & 62.53 & 70.23 & 75.47 & 63.54 & 59.05 & 63.66 & 66.22 & 49.14 & 52.95 & \textcolor{gray}{20.79} \\
\textbf{Mistral Large}\raisebox{-0.04in} & 60.84 & 68.76 & 72.32 & 56.83 & 61.21 & 55.35 & \textbf{70.17} & 43.44 & 58.66 & \textcolor{gray}{24.47} \\
\textbf{GPT-3.5} & 60.32 & 65.38 & 78.42 & 64.56 & 54.89 & 67.89 & 60.11 & 41.42 & 49.90 & \textcolor{gray}{29.99} \\
\textbf{Gemini 1.0 Pro} & 54.06 & 56.04 & 72.35 & 56.35 & 47.70 & 50.56 & 61.02 & 35.89 & 52.55 & \textcolor{gray}{25.69} \\
\textbf{Llama3-70b-Instruct} & 52.92 & 59.67 & 67.09 & 57.14 & 43.10 & 55.49 & 58.21 & 41.67 & 40.94 & \textcolor{gray}{--} \\
\bottomrule
\end{tabular}
}
\label{leaderboard}
\centering
\caption{Open-LLM Leaderboard for small-scale model regime.}
\resizebox{\textwidth}{!}{%
\begin{tabular}{l|c|cccccccc}
\toprule
\textbf{Model} & \textbf{Overall} & \textbf{MMLU} & \textbf{ARC} & \textbf{WG} & \textbf{PIQA} & \textbf{CSQA} & \textbf{Race} & \textbf{MedMCQA} & \textbf{OBQA} \\
\midrule
Qwen1.5 (1.8B)                     & \bf 21.68 & 9.99    & 15.84 &\bf 40.96 &\bf 15.52 &\bf 31.13 & \bf 34.91 & \bf 4.70 & \bf 20.37   \\
Gemma (2B)                      & 16.66 & \bf 17.52 & \bf 23.93 & 16.10 & 15.09 & 27.46 & 14.32 & 4.57 & 14.26  \\
SlimPajama-DC (1.3B)                     & 9.60  & 9.22  & 14.95 & 14.76 & 5.32  & 9.01  & 16.19 & 1.68 & 5.70  \\
RedPajama (1.3B)                  & 9.00  & 9.21  & 13.50 & 16.97 & 0.86  & 11.41 & 14.35 & 1.86 & 3.87  \\
OLMo (1.2B)                       & 8.85  & 8.54  & 13.18 & 6.16  & 8.05  & 13.10  & 13.61 & 2.07 & 6.11   \\
Pythia (1.4B)                   & 8.79  & 9.66  & 14.69 & 11.52 & 4.17  & 9.01  & 12.76 & 3.19 & 5.30  \\
TinyLlama (1.1B)                & 8.45  & 8.94  & 13.31 & 12.23 & 3.59  & 6.06  & 16.7  & 2.07 & 4.68  \\
OPT (1.3B)                      & 7.89  & 7.40  & 11.83 & 12.47 & 4.48  & 7.61  & 13.61 & 1.25 & 4.48  \\
GPT-Neo (1.3B)                  & 7.42  & 6.94  & 9.69  & 10.81 & 4.31  & 6.34  & 13.75 & 2.63 & 4.89  \\
Cerebras-GPT (1.3B)             & 4.86  & 5.37  & 4.43  & 9.31  & 2.16  & 6.20   & 6.90   & 1.04 & 3.46  \\
\bottomrule
\end{tabular}
}
\label{small_leaderboard}
\end{table}

\vspace{-0.1in}
\subsection{Leaderboard and Arena} 
\vspace{-0.05in}
The overall ranking of models for our benchmark is represented in Table \ref{leaderboard} and Table \ref{small_leaderboard}. 
The performance of GPT-4o overall demonstrates its leading edge, with an accuracy of 70.15\%, which indicates its robustness in open-style question answering tasks compared to other models. It is followed by GPT-4-1106-preview with 65.93\%, and Claude-3 Opus with 62.68\%. These results highlight the advanced capabilities of the GPT-4 series. Mid-tier models like Mistral Large and GPT-3.5 perform well but are not on par with the top performers. On the other hand, models like Gemini 1.0 Pro and Llama3-70b-Instruct lag behind in terms of the capabilities to answer the open-style questions.

The performance evaluation of smaller-scale LLMs reveals that Qwen1.5 leads with an overall accuracy of 21.68\%, significantly outperforming the other models in this category. Gemma follows with 16.66\%, indicating a considerable gap in performance compared to the top model. The remaining models score below 10.00\%, highlighting their limited abilities to answer the open-style questions. Almost all of the models struggle significantly with questions from MedMCQA dataset, showing an accuracy below of 5\%.

\vspace{-0.05in}
\section{Conclusion}
\vspace{-0.05in}
We proposed {\texttt{Open-LLM-Leaderboard}} for LLM evaluation and comprehensively examined its efficacy using open-style questions from nine datasets on OSQ-bench. Different from previous works that rely on human evaluation or thousands of crowd users on Chatbot Arena, we can have a benchmark for chat LLMs in a fast, automatic, and cheap scheme. Our results show a highly correlated level of agreement with humans, indicating a foundation for an LLM-based evaluation benchmark and framework using open-style questions.

\section*{Limitations and Ethics Statement} \label{limit}

We have discussed multiple advantages of employing open-style questions over multiple-choice questions used in prior works. However, the LLM Leaderboard, as a tool for evaluating and benchmarking LLMs, has several common limitations itself. Firstly, the performance metrics used may not fully capture the nuanced capabilities of each model, especially in areas that require an understanding of context, creativity, or common sense reasoning. Secondly, the benchmark datasets may not be comprehensive enough to cover all possible domains and scenarios, leading to a potential bias towards certain types of questions or tasks. Thirdly, due to the rapidly evolving nature of the field, models may quickly become outdated, meaning the leaderboard may not always reflect the most current state of the art. Since our benchmark utilizes public datasets and our corpus consists of questions and answers, user privacy concerns are minimal.

\bibliographystyle{plain}
\bibliography{neurips_data_2024}

\begin{thebibliography}{10}

\bibitem{llama3modelcard}
AI@Meta.
\newblock Llama 3 model card.
\newblock 2024.

\bibitem{claude}
Anthropic.
\newblock Model card and evaluations for claude models, 2023.

\bibitem{claude_chat}
{Anthropic}.
\newblock \url{https://www.anthropic.com/claude}, 2024.

\bibitem{qwen}
Jinze Bai, Shuai Bai, Yunfei Chu, Zeyu Cui, Kai Dang, Xiaodong Deng, Yang Fan,
  Wenbin Ge, Yu~Han, Fei Huang, Binyuan Hui, Luo Ji, Mei Li, Junyang Lin, Runji
  Lin, Dayiheng Liu, Gao Liu, Chengqiang Lu, Keming Lu, Jianxin Ma, Rui Men,
  Xingzhang Ren, Xuancheng Ren, Chuanqi Tan, Sinan Tan, Jianhong Tu, Peng Wang,
  Shijie Wang, Wei Wang, Shengguang Wu, Benfeng Xu, Jin Xu, An~Yang, Hao Yang,
  Jian Yang, Shusheng Yang, Yang Yao, Bowen Yu, Hongyi Yuan, Zheng Yuan,
  Jianwei Zhang, Xingxuan Zhang, Yichang Zhang, Zhenru Zhang, Chang Zhou,
  Jingren Zhou, Xiaohuan Zhou, and Tianhang Zhu.
\newblock Qwen technical report.
\newblock {\em arXiv preprint arXiv:2309.16609}, 2023.

\bibitem{open-llm-leaderboard}
Edward Beeching, Clémentine Fourrier, Nathan Habib, Sheon Han, Nathan Lambert,
  Nazneen Rajani, Omar Sanseviero, Lewis Tunstall, and Thomas Wolf.
\newblock Open llm leaderboard.
\newblock
  \url{https://huggingface.co/spaces/HuggingFaceH4/open_llm_leaderboard}, 2023.

\bibitem{biderman2023pythia}
Stella Biderman, Hailey Schoelkopf, Quentin~Gregory Anthony, Herbie Bradley,
  Kyle O’Brien, Eric Hallahan, Mohammad~Aflah Khan, Shivanshu Purohit,
  USVSN~Sai Prashanth, Edward Raff, et~al.
\newblock Pythia: A suite for analyzing large language models across training
  and scaling.
\newblock In {\em International Conference on Machine Learning}, pages
  2397--2430. PMLR, 2023.

\bibitem{Bisk_Zellers_Le}
Yonatan Bisk, Rowan Zellers, Ronan Le~bras, Jianfeng Gao, and Yejin Choi.
\newblock Piqa: Reasoning about physical commonsense in natural language.
\newblock {\em Proceedings of the AAAI Conference on Artificial Intelligence},
  34(05):7432--7439, Apr. 2020.

\bibitem{gpt-neo}
Sid Black, Gao Leo, Phil Wang, Connor Leahy, and Stella Biderman.
\newblock {GPT-Neo: Large Scale Autoregressive Language Modeling with
  Mesh-Tensorflow}, March 2021.
\newblock {If you use this software, please cite it using these metadata.}

\bibitem{NEURIPS2020_1457c0d6}
Tom Brown, Benjamin Mann, Nick Ryder, Melanie Subbiah, Jared~D Kaplan, Prafulla
  Dhariwal, Arvind Neelakantan, Pranav Shyam, Girish Sastry, Amanda Askell,
  Sandhini Agarwal, et~al.
\newblock Language models are few-shot learners.
\newblock In H.~Larochelle, M.~Ranzato, R.~Hadsell, M.F. Balcan, and H.~Lin,
  editors, {\em Advances in Neural Information Processing Systems}, volume~33,
  pages 1877--1901. Curran Associates, Inc., 2020.

\bibitem{vicuna2023}
Wei-Lin Chiang, Zhuohan Li, Zi~Lin, Ying Sheng, Zhanghao Wu, Hao Zhang, Lianmin
  Zheng, Siyuan Zhuang, Yonghao Zhuang, Joseph~E. Gonzalez, Ion Stoica, and
  Eric~P. Xing.
\newblock Vicuna: An open-source chatbot impressing gpt-4 with 90\%* chatgpt
  quality, March 2023.

\bibitem{Chung_2023}
John Chung, Ece Kamar, and Saleema Amershi.
\newblock Increasing diversity while maintaining accuracy: Text data generation
  with large language models and human interventions.
\newblock In {\em Proceedings of the 61st Annual Meeting of the Association for
  Computational Linguistics (Volume 1: Long Papers)}, 2023.

\bibitem{Clark2018ThinkYH}
Peter Clark, Isaac Cowhey, Oren Etzioni, Tushar Khot, Ashish Sabharwal, Carissa
  Schoenick, and Oyvind Tafjord.
\newblock Think you have solved question answering? try arc, the ai2 reasoning
  challenge.
\newblock {\em ArXiv}, abs/1803.05457, 2018.

\bibitem{doi:10.1177/001316446002000104}
Jacob Cohen.
\newblock A coefficient of agreement for nominal scales.
\newblock {\em Educational and Psychological Measurement}, 20(1):37--46, 1960.

\bibitem{dey2023cerebrasgpt}
Nolan Dey, Gurpreet Gosal, Zhiming, Chen, Hemant Khachane, William Marshall,
  Ribhu Pathria, Marvin Tom, and Joel Hestness.
\newblock Cerebras-gpt: Open compute-optimal language models trained on the
  cerebras wafer-scale cluster.
\newblock {\em arXiv preprint:2304.03208}, 2023.

\bibitem{eval-harness}
Leo Gao, Jonathan Tow, Baber Abbasi, Stella Biderman, Sid Black, Anthony
  DiPofi, Charles Foster, Laurence Golding, Jeffrey Hsu, Alain Le~Noac'h,
  Haonan Li, Kyle McDonell, Niklas Muennighoff, Chris Ociepa, Jason Phang,
  Laria Reynolds, Hailey Schoelkopf, Aviya Skowron, Lintang Sutawika, Eric
  Tang, Anish Thite, Ben Wang, Kevin Wang, and Andy Zou.
\newblock A framework for few-shot language model evaluation, 12 2023.

\bibitem{gemini_pro}
{Google}.
\newblock \url{https://ai.google.dev/}, 2023.

\bibitem{Groeneveld2023OLMo}
Dirk Groeneveld, Iz~Beltagy, Pete Walsh, Akshita Bhagia, Rodney Kinney, Oyvind
  Tafjord, Ananya~Harsh Jha, Hamish Ivison, Ian Magnusson, Yizhong Wang, Shane
  Arora, David Atkinson, Russell Authur, Khyathi Chandu, Arman Cohan, Jennifer
  Dumas, Yanai Elazar, Yuling Gu, Jack Hessel, Tushar Khot, William Merrill,
  Jacob Morrison, Niklas Muennighoff, Aakanksha Naik, Crystal Nam, Matthew~E.
  Peters, Valentina Pyatkin, Abhilasha Ravichander, Dustin Schwenk, Saurabh
  Shah, Will Smith, Nishant Subramani, Mitchell Wortsman, Pradeep Dasigi,
  Nathan Lambert, Kyle Richardson, Jesse Dodge, Kyle Lo, Luca Soldaini, Noah~A.
  Smith, and Hannaneh Hajishirzi.
\newblock Olmo: Accelerating the science of language models.
\newblock {\em Preprint}, 2024.

\bibitem{hendryckstest2021}
Dan Hendrycks, Collin Burns, Steven Basart, Andy Zou, Mantas Mazeika, Dawn
  Song, and Jacob Steinhardt.
\newblock Measuring massive multitask language understanding.
\newblock {\em Proceedings of the International Conference on Learning
  Representations (ICLR)}, 2021.

\bibitem{lai-etal-2017-race}
Guokun Lai, Qizhe Xie, Hanxiao Liu, Yiming Yang, and Eduard Hovy.
\newblock {RACE}: Large-scale {R}e{A}ding comprehension dataset from
  examinations.
\newblock In {\em Proceedings of the 2017 Conference on Empirical Methods in
  Natural Language Processing}, 2017.

\bibitem{li2023starcoder}
Raymond Li, Loubna~Ben Allal, Yangtian Zi, Niklas Muennighoff, Denis Kocetkov,
  Chenghao Mou, Marc Marone, Christopher Akiki, et~al.
\newblock Starcoder: may the source be with you!
\newblock {\em arXiv preprint arXiv:23.05.061161}, 2023.

\bibitem{alpaca_eval}
Xuechen Li, Tianyi Zhang, Yann Dubois, Rohan Taori, Ishaan Gulrajani, Carlos
  Guestrin, Percy Liang, and Tatsunori~B. Hashimoto.
\newblock Alpacaeval: An automatic evaluator of instruction-following models.
\newblock \url{https://github.com/tatsu-lab/alpaca_eval}, 2023.

\bibitem{liu2023learning}
Yixin Liu, Kejian Shi, Katherine~S He, Longtian Ye, Alexander~R. Fabbri,
  Pengfei Liu, Dragomir Radev, and Arman Cohan.
\newblock On learning to summarize with large language models as references.
\newblock {\em arXiv preprint arXiv:2305.14239}, 2023.

\bibitem{mihaylov-etal-2018-suit}
Todor Mihaylov, Peter Clark, Tushar Khot, and Ashish Sabharwal.
\newblock Can a suit of armor conduct electricity? a new dataset for open book
  question answering.
\newblock In {\em Proceedings of the 2018 Conference on Empirical Methods in
  Natural Language Processing}, 2018.

\bibitem{mistral_chat}
{Mistral}.
\newblock \url{https://chat.mistral.ai/chat}, 2024.

\bibitem{mosaicml_mpt1b_redpajama_200b}
{MosaicML}.
\newblock Mpt-1b redpajama-200b.
\newblock \url{https://huggingface.co/mosaicml/mpt-1b-redpajama-200b}.
\newblock Accessed: 2024-04-29.

\bibitem{openai_evals}
{OpenAI}.
\newblock Openai evals.
\newblock \url{https://github.com/openai/evals}.

\bibitem{openai_chat}
{OpenAI}.
\newblock \url{https://chat.openai.com/chat}, 2022.

\bibitem{openai2024gpt4}
OpenAI.
\newblock Gpt-4 technical report.
\newblock {\em arxiv preprint arXiv:2303.08774}, 2024.

\bibitem{pmlr-v174-pal22a}
Ankit Pal, Logesh~Kumar Umapathi, and Malaikannan Sankarasubbu.
\newblock Medmcqa: A large-scale multi-subject multi-choice dataset for medical
  domain question answering.
\newblock In {\em Proceedings of the Conference on Health, Inference, and
  Learning}, 2022.

\bibitem{pezeshkpour2023large}
Pouya Pezeshkpour and Estevam Hruschka.
\newblock Large language models sensitivity to the order of options in
  multiple-choice questions.
\newblock {\em arXiv preprint arXiv:2308.11483}, 2023.

\bibitem{Robinson2022LeveragingLL}
Joshua Robinson, Christopher Rytting, and David Wingate.
\newblock Leveraging large language models for multiple choice question
  answering.
\newblock {\em ArXiv}, abs/2210.12353, 2022.

\bibitem{robinson2023leveraging}
Joshua Robinson, Christopher~Michael Rytting, and David Wingate.
\newblock Leveraging large language models for multiple choice question
  answering.
\newblock {\em arXiv preprint arXiv:2210.12353}, 2023.

\bibitem{r2024code}
Baptiste Rozière, Jonas Gehring, Fabian Gloeckle, Sten Sootla, Itai Gat,
  Xiaoqing~Ellen Tan, Yossi Adi, Jingyu Liu, Romain Sauvestre, Tal Remez,
  Jérémy Rapin, Artyom Kozhevnikov, Ivan Evtimov, Joanna Bitton, Manish
  Bhatt, Cristian~Canton Ferrer, Aaron Grattafiori, Wenhan Xiong, Alexandre
  Défossez, Jade Copet, Faisal Azhar, Hugo Touvron, Louis Martin, Nicolas
  Usunier, Thomas Scialom, and Gabriel Synnaeve.
\newblock Code llama: Open foundation models for code.
\newblock {\em arXiv preprint arXiv:2308.12950}, 2024.

\bibitem{10.1145/3474381}
Keisuke Sakaguchi, Ronan~Le Bras, Chandra Bhagavatula, and Yejin Choi.
\newblock Winogrande: an adversarial winograd schema challenge at scale.
\newblock {\em Commun. ACM}, 64(9):99–106, aug 2021.

\bibitem{shen2023slimpajama}
Zhiqiang Shen, Tianhua Tao, Liqun Ma, Willie Neiswanger, Joel Hestness, Natalia
  Vassilieva, Daria Soboleva, and Eric Xing.
\newblock Slimpajama-dc: Understanding data combinations for llm training.
\newblock {\em arXiv preprint arXiv:2309.10818}, 2023.

\bibitem{talmor-etal-2019-commonsenseqa}
Alon Talmor, Jonathan Herzig, Nicholas Lourie, and Jonathan Berant.
\newblock {C}ommonsense{QA}: A question answering challenge targeting
  commonsense knowledge.
\newblock In {\em Proceedings of the 2019 Conference of the North {A}merican
  Chapter of the Association for Computational Linguistics: Human Language
  Technologies, Volume 1 (Long and Short Papers)}, 2019.

\bibitem{alpaca}
Rohan Taori, Ishaan Gulrajani, Tianyi Zhang, Yann Dubois, Xuechen Li, Carlos
  Guestrin, Percy Liang, and Tatsunori~B. Hashimoto.
\newblock Stanford alpaca: An instruction-following llama model.
\newblock \url{https://github.com/tatsu-lab/stanford_alpaca}, 2023.

\bibitem{geminiteam2023gemini}
Gemini Team.
\newblock Gemini: A family of highly capable multimodal models.
\newblock {\em arXiv preprint arXiv:2312.11805}, 2023.

\bibitem{gemmateam2024gemma}
Gemma Team.
\newblock Gemma: Open models based on gemini research and technology.
\newblock {\em arXiv preprint arXiv:2403.08295}, 2024.

\bibitem{touvron2023llama}
Hugo Touvron, Thibaut Lavril, Gautier Izacard, Xavier Martinet, Marie-Anne
  Lachaux, Timothée Lacroix, Baptiste Rozière, Naman Goyal, Eric Hambro,
  Faisal Azhar, et~al.
\newblock Llama: Open and efficient foundation language models.
\newblock {\em arXiv preprint arXiv:2302.13971}, 2023.

\bibitem{touvron2023llama2}
Hugo Touvron, Louis Martin, Kevin Stone, Peter Albert, Amjad Almahairi, Yasmine
  Babaei, Nikolay Bashlykov, Soumya Batra, Prajjwal Bhargava, Shruti Bhosale,
  et~al.
\newblock Llama 2: Open foundation and fine-tuned chat models.
\newblock {\em arXiv preprint arXiv:2307.09288}, 2023.

\bibitem{wang2023large}
Peiyi Wang, Lei Li, Liang Chen, Zefan Cai, Dawei Zhu, Binghuai Lin, Yunbo Cao,
  Qi~Liu, Tianyu Liu, and Zhifang Sui.
\newblock Large language models are not fair evaluators.
\newblock {\em arXiv preprint arXiv:2305.17926}, 2023.

\bibitem{DBLP:journals/corr/abs-1910-03771}
Thomas Wolf, Lysandre Debut, Victor Sanh, Julien Chaumond, Clement Delangue,
  Anthony Moi, Pierric Cistac, Tim Rault, R{\'{e}}mi Louf, Morgan Funtowicz,
  and Jamie Brew.
\newblock Huggingface's transformers: State-of-the-art natural language
  processing.
\newblock {\em CoRR}, abs/1910.03771, 2019.

\bibitem{zellers-etal-2019-hellaswag}
Rowan Zellers, Ari Holtzman, Yonatan Bisk, Ali Farhadi, and Yejin Choi.
\newblock {H}ella{S}wag: Can a machine really finish your sentence?
\newblock In {\em Proceedings of the 57th Annual Meeting of the Association for
  Computational Linguistics}, 2019.

\bibitem{pmlr-v202-zhang23m}
Biao Zhang, Barry Haddow, and Alexandra Birch.
\newblock Prompting large language model for machine translation: A case study.
\newblock In {\em Proceedings of the 40th International Conference on Machine
  Learning}, 2023.

\bibitem{zhang2024tinyllama}
Peiyuan Zhang, Guangtao Zeng, Tianduo Wang, and Wei Lu.
\newblock Tinyllama: An open-source small language model.
\newblock {\em arXiv preprint:2401.02385}, 2024.

\bibitem{zhang2022opt}
Susan Zhang, Stephen Roller, Naman Goyal, Mikel Artetxe, Moya Chen, Shuohui
  Chen, Christopher Dewan, Mona Diab, Xian Li, Xi~Victoria Lin, Todor Mihaylov,
  Myle Ott, Sam Shleifer, Kurt Shuster, Daniel Simig, Punit~Singh Koura, Anjali
  Sridhar, Tianlu Wang, and Luke Zettlemoyer.
\newblock Opt: Open pre-trained transformer language models.
\newblock {\em arXiv preprint:2205.01068}, 2022.

\bibitem{zheng2024large}
Chujie Zheng, Hao Zhou, Fandong Meng, Jie Zhou, and Minlie Huang.
\newblock Large language models are not robust multiple choice selectors.
\newblock {\em arXiv preprint arXiv:2309.03882}, 2024.

\bibitem{zheng2023judging}
Lianmin Zheng, Wei-Lin Chiang, Ying Sheng, Siyuan Zhuang, Zhanghao Wu, Yonghao
  Zhuang, Zi~Lin, Zhuohan Li, Dacheng Li, Eric~P. Xing, Hao Zhang, Joseph~E.
  Gonzalez, and Ion Stoica.
\newblock Judging llm-as-a-judge with mt-bench and chatbot arena.
\newblock {\em arXiv preprint arXiv:2306.05685}, 2023.

\bibitem{zhu2023multilingual}
Wenhao Zhu, Hongyi Liu, Qingxiu Dong, Jingjing Xu, Shujian Huang, Lingpeng
  Kong, Jiajun Chen, and Lei Li.
\newblock Multilingual machine translation with large language models:
  Empirical results and analysis.
\newblock {\em arXiv preprint 2304.04675}, 2023.

\end{thebibliography}


\newpage

\appendix

\section*{\Large{Appendix}}

\section{Reproducibility Statement}

We will make all our filtered open-style data (MMLU, ARC, HellaSwag, WinoGrande, PIQA, CommonsenseQA, Race, MedMCQA, and OpenbookQA) used in our experiments of Sec.~\ref{experiments} and preprocessing scripts publicly available. Detailed data statistics are provided in Sec.~\ref{stats}. Considering the potential high costs associated with gathering and reproducing our LLM response data from the ground up, we will make available all responses from the various LLMs and their corresponding evaluation results to support and simplify the reproducibility of our work. The OpenAI APIs we used include gpt-3.5-turbo-1106, gpt-4.0-1106-preview, gpt-4o (for response collection), and gpt-4.0-0125-preview (for filtering and post-evaluation); Claude 3: claude-3-opus-20240229; Gemini-Pro: gemini-pro, and Mistral: mistral-large-latest.

\section{More Results on Gemini Pro and Stage1 Filtering}
\vspace{-0.02in}

\begin{figure}[h]
    \centering
    \vspace{-0.2in}
    \includegraphics[width=0.95\linewidth]{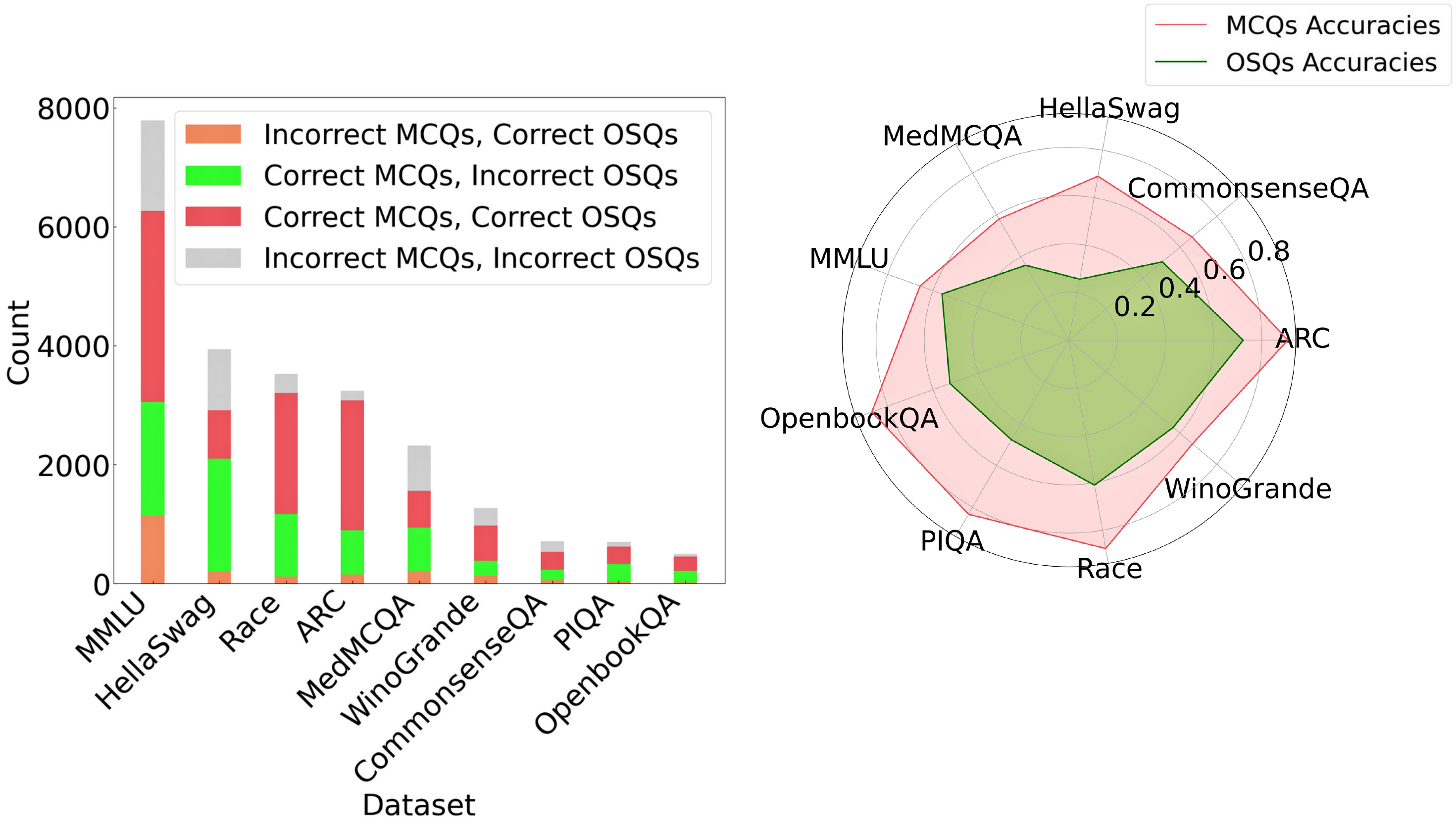}
  \vspace{-0.08in}
  \caption{Performance comparison of Gemini Pro on multiple-choice and open-style response questions across diverse datasets, as shown by the count of correct and incorrect answers in the left bar chart and model accuracy in the right radar chart. }
  \label{TF-g}
  \vspace{-0.14in}
\end{figure}\

\begin{figure}[h]
    \centering
    \includegraphics[width=0.75\linewidth]{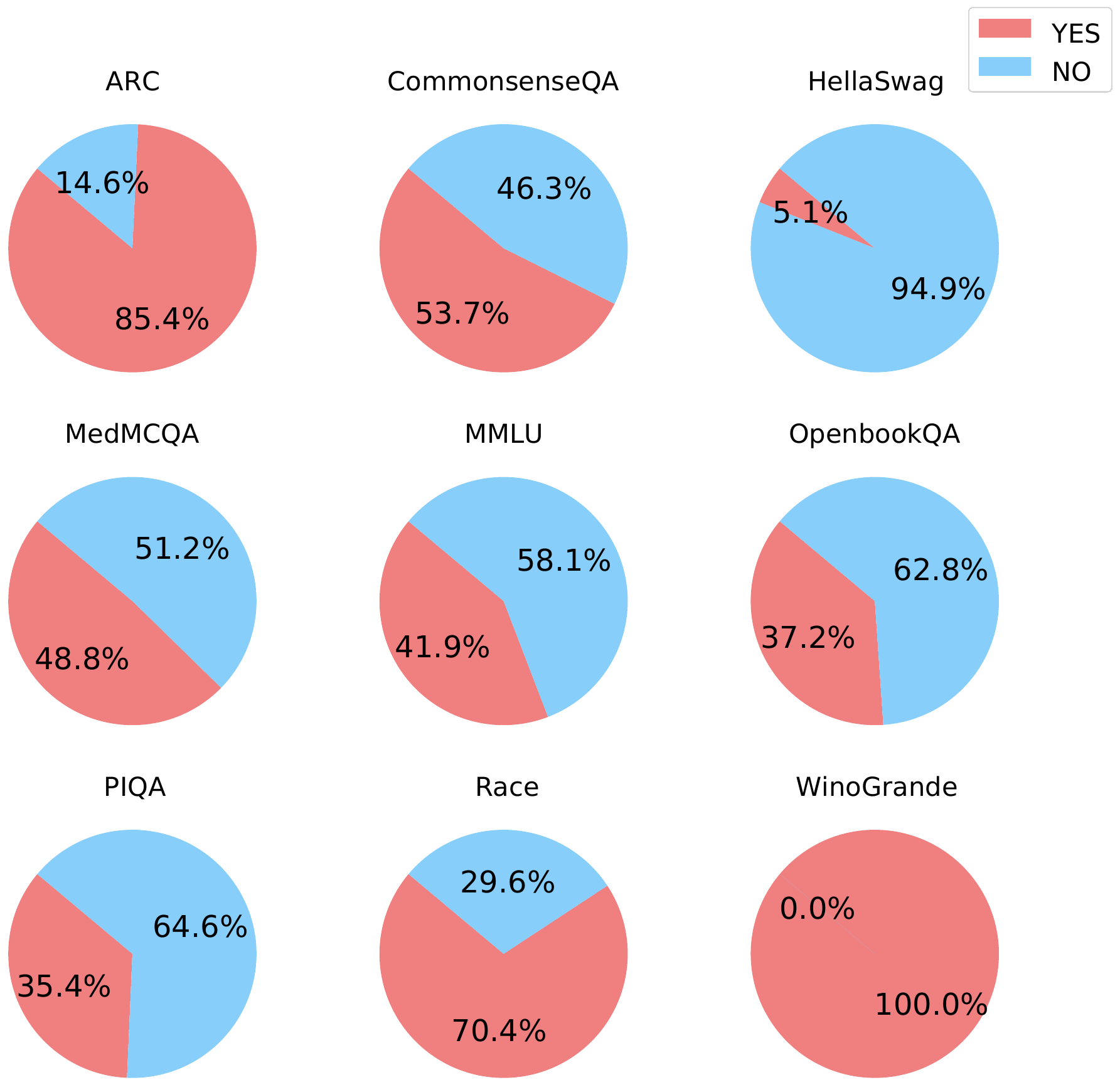}
  \vspace{-0.1in}
  \caption{Initial filtering results across datasets, showcasing the percentages of questions that are suitable (Yes) for open-style formatting and those that are not (No).}
  \label{before}
  \vspace{-0.1in}
\end{figure}

\vspace{-0.15in}
\section{More Sample Questions}
\vspace{-0.05in}
We provide more sample questions that are correct for open-style and incorrect for multiple choice in Table~\ref{more_sample_1}, and vice versa in Table~\ref{more_sample_2}. These examples demonstrate how the model's ability varies in handling different question formats.

\begin{table}[h]
\vspace{-2mm}
 \caption{Example of an Incorrect MCQ and a Correct OSQ in MMLU.}
\centering
    \scalebox{0.90}{
    \begin{tabular}[c]{@{}p{16cm}@{}}
      \toprule
      \texttt{Question that is correct for OSQ and incorrect for MCQ: \textcolor{red}{How many CO2 and ATP molecules are formed during one complete turn of the tricarboxylic acid cycle (Krebs' cycle)?}} \\
      \texttt{Options:} 
      \texttt{A. \textcolor{blue}{2CO2 and 2ATP} \ \  B. \textcolor{blue}{2CO2 and 16ATP}} \ \
      \texttt{C. \textcolor{blue}{2CO2 and 12ATP} \ \  D. \textcolor{blue}{2CO2 and 1ATP}} \\
      \texttt{Ground Truth: \textcolor{red}{ \underline{D}}} \\ 
       \texttt{MCQ\_Answer: \textcolor{red}{ \underline{A}}} \\ 
      
      \texttt{OSQ\_Answer: \textcolor{red}{During one complete turn of the tricarboxylic acid cycle (also known as the Krebs' cycle or the citric acid cycle), a series of chemical reactions occur in the mitochondria of cells that result in the production of carbon dioxide (CO2) and adenosine triphosphate (ATP), among other molecules.}} \\
     \textcolor{red}{\texttt{\hspace{0.5cm}For each acetyl-CoA molecule that enters the cycle, the following are produced:}} \\
      \textcolor{red}{\texttt{\hspace{0.5cm}- 2 molecules of CO2: Each acetyl-CoA contributes two carbons to the cycle, which are eventually released as two molecules of carbon dioxide.}} \\
      \textcolor{red}{\texttt{\hspace{0.5cm}- 1 molecule of ATP: One molecule of ATP (or an equivalent, such as GTP, depending on the cell type) is produced directly in the cycle through substrate-level phosphorylation.}} \\
      \textcolor{red}{\texttt{\hspace{0.5cm}However, the cycle also produces reduced coenzymes, specifically NADH and FADH2, which are used in the electron transport chain to produce additional ATP through oxidative phosphorylation. For each acetyl-CoA that enters the cycle:}} \\
      \textcolor{red}{\texttt{\hspace{0.5cm}- 3 molecules of NADH are produced, which can be used to generate approximately 2.5 ATP each through the electron transport chain.}} \\
      \textcolor{red}{\texttt{\hspace{0.5cm}- 1 molecule of FADH2 is produced, which can be used to generate approximately 1.5 ATP through the electron transport chain.}} \\
      \textcolor{red}{\texttt{\hspace{0.5cm}Therefore, while only 1 ATP is produced directly in the cycle per acetyl-CoA, the NADH and FADH2 generated can lead to the production of additional ATP in the electron transport chain. The exact number of ATP molecules produced can vary depending on the efficiency of the electron transport chain and the ATP synthase complex in different organisms and cell types.}} \\
      
    \bottomrule
    \end{tabular}
    \vspace{-3.6mm}
    }
    \label{more_sample_1}
    \vspace{-3.5mm}
\end{table}

\begin{table}[t]

\centering
    \caption{Example of a Correct MCQ and an Incorrect OSQ in MMLU.}
    \scalebox{0.90}{
    \begin{tabular}[c]{@{}p{16cm}@{}}
      \toprule
      \texttt{Question that is correct for MCQ and incorrect for OSQ: \textcolor{red}{A tree is a connected graph with no cycles. How many nonisomorphic trees with 5 vertices exist?}} \\
      \texttt{Options:} 
      \texttt{A. \textcolor{blue}{1} \ \  B. \textcolor{blue}{2}} \ \
      \texttt{C. \textcolor{blue}{3} \ \  D. \textcolor{blue}{4}} \\
      \texttt{Ground Truth: \textcolor{red}{ \underline{C}}} \\ 
       \texttt{MCQ\_Answer: \textcolor{red}{ \underline{C}}} \\ 
      
     \texttt{OSQ\_Answer: \textcolor{red}{\texttt{To determine the number of nonisomorphic trees with 5 vertices, we can use Cayley's formula, which states that the number of trees on \( n \) labeled vertices is \( n^{n-2} \). However, this formula gives the number of trees when vertices are labeled, meaning that it counts different labelings of the same tree shape as distinct. Since we are interested in nonisomorphic trees (which do not consider different labelings as distinct), we cannot directly use Cayley's formula.}}} \\
      \textcolor{red}{\texttt{Instead, we need to enumerate the trees by considering their possible structures. For a tree with 5 vertices, we can have the following configurations:}} \\
      \textcolor{red}{\texttt{1. A linear chain (path) where each vertex is connected to two others except for the two end vertices, which are connected to only one other vertex. This is one possible tree.}} \\
      \textcolor{red}{\texttt{2. A "star" tree where one central vertex is connected to the other four vertices. This is another possible tree.}} \\
      \textcolor{red}{\texttt{3. A tree where one vertex is connected to three others, and one of those three is connected to the fifth vertex. This structure can be seen as a "T" shape with an extra branch on one of the endpoints. There are three nonisomorphic trees of this type because the branch can be on any of the three endpoints.}} \\
      \textcolor{red}{\texttt{4. A tree where two vertices are connected to a central vertex, and each of those two vertices is connected to one other vertex (creating a structure with a central vertex, two intermediate vertices, and two leaf vertices). This structure can be seen as a "Y" shape. There is only one nonisomorphic tree of this type because any permutation of the leaf vertices does not create a fundamentally different tree structure.}} \\
      \textcolor{red}{\texttt{Adding these up, we have:}} \\
      \textcolor{red}{\texttt{1 (linear chain) + 1 (star) + 3 (T-shaped with an extra branch) + 1 (Y-shaped) = 6 nonisomorphic trees with 5 vertices.}} \\
      
    \bottomrule
    \end{tabular}
    \vspace{-3.6mm}
    }
    \label{more_sample_2}
    \vspace{-4.3mm}
\end{table}

\section{Prompts for Different Datasets}\label{pfdd}
The following are examples of prompts for different datasets. They allow models to better understand and respond to specific contexts and objectives in each dataset.
\begin{table}[h]
\centering
    \caption{Prompt for MMLU dataset.}
    \scalebox{0.90}{
    \begin{tabular}[c]{@{}p{16.5cm}@{}}
      \toprule
      \textbf{MCQ Prompt:} \texttt{The following is the multiple choice question about abstract algebra. Please select the correct answer from the options A, B, C, D. For example, if you think the correct answer is A, your response should be 'A'.}\\
      \texttt{Question: Find the degree for the given field extension Q(sqrt(2), sqrt(3), sqrt(18)) over Q.}\\
      \texttt{A. \textcolor{blue}{0}}\\
      \texttt{B. \textcolor{blue}{4}}\\
      \texttt{C. \textcolor{blue}{2}}\\
      \texttt{D. \textcolor{blue}{6}}\\
    \texttt{Answer:}
    \\
    \\
    \textbf{Open-Style Prompt:}
    \texttt{Answer the following question about abstract algebra.}\\
    \texttt{Question: Find the degree for the given field extension Q(sqrt(2), sqrt(3), sqrt(18)) over Q.}\\
    \texttt{Answer:} \\
    \bottomrule
    \end{tabular}
    \vspace{-3.6mm}
    }
    \label{prompt_mmlu}
    \vspace{-4.3mm}
\end{table}

\begin{table}[h]
\centering
    \caption{Prompt for ARC dataset.}
    \scalebox{0.90}{
    \begin{tabular}[c]{@{}p{16.5cm}@{}}
      \toprule
      
      \textbf{MCQ Prompt:} \texttt{The following is the multiple choice question. Please select the correct answer from the options A, B, C, D. For example, if you think the correct answer is A, your response should be 'A'.}\\
      \texttt{Question: Tiny organisms called plankton live in oceans. Some plankton can take energy from the Sun and turn it into food. How are plankton most useful to the animals that live in the ocean?}\\
      \texttt{A. \textcolor{blue}{Plankton are colorful.}}\\
      \texttt{B. \textcolor{blue}{Plankton clean the water.}}\\
      \texttt{C. \textcolor{blue}{Plankton release oxygen.}}\\
      \texttt{D. \textcolor{blue}{Plankton reproduce quickly.}}\\
    \texttt{Answer:}
    \\
    \\
    \textbf{Open-Style Prompt:}
    Answer the following question.\\
    \texttt{Question: Tiny organisms called plankton live in oceans. Some plankton can take energy from the Sun and turn it into food. How are plankton most useful to the animals that live in the ocean?}\\
    \texttt{Answer:}\\
    \bottomrule
    \end{tabular}
    \vspace{-3.6mm}
    }
    \label{prompt_arc}
    \vspace{-4.3mm}
\end{table}

\begin{table}[h]
\centering
    \caption{Prompt for CommonsenseQA dataset.}
    \scalebox{0.90}{
    \begin{tabular}[c]{@{}p{16.5cm}@{}}
      \toprule
      \textbf{MCQ Prompt:}
     \texttt{The following is the multiple choice question. Please select the correct answer from the options A, B, C, D, E. For example, if you think the correct answer is A, your response should be 'A'.}\\
      \texttt{Question: A revolving door is convenient for two direction travel, but it also serves as a security measure at a what?}\\
      \texttt{A. \textcolor{blue}{bank}}\\
      \texttt{B. \textcolor{blue}{library}}\\
      \texttt{C. \textcolor{blue}{department store}}\\
      \texttt{D. \textcolor{blue}{mall}}\\
      \texttt{E. \textcolor{blue}{New York}}\\
    \texttt{Answer:}
    \\
    \\
    \textbf{Open-Style Prompt:} \texttt{You will be presented with a variety of questions that require an understanding of everyday scenarios, human behaviors, and common sense. Your task is to provide the best possible answer to each question based solely on your understanding and reasoning.}\\
    \texttt{Question: A revolving door is convenient for two direction travel, but it also serves as a security measure at a what?}\\
    \texttt{Answer:}\\
    \bottomrule
    \end{tabular}
    \vspace{-3.6mm}
    }

    \label{prompt_csqa}
    \vspace{-4.3mm}
\end{table}

\begin{table}[h]
\centering
    \caption{Prompt for MedMCQA dataset.}
    \scalebox{0.95}{
    \begin{tabular}[c]{@{}p{16.5cm}@{}}
      \toprule
      \textbf{MCQ Prompt:}
     \texttt{The following is the multiple choice question about medicine. Please select the correct answer from the options A, B, C, D. For example, if you think the correct answer is A, your response should be 'A'.}\\
      \texttt{Question: Modulus of elasticity means:}\\
      \texttt{A. \textcolor{blue}{Rigidity or stiffness of the material}}\\
      \texttt{B. \textcolor{blue}{Ability to be stretched with permanent deformation}}\\
      \texttt{C. \textcolor{blue}{Ductility of a material}}\\
      \texttt{D. \textcolor{blue}{Malleability of the metal}}\\
    \texttt{Answer:}
    \\
    \\
    \textbf{Open-Style Prompt:} \texttt{Answer the following question about medicine.}\\
    \texttt{Question: Modulus of elasticity means:}\\
    \texttt{Answer:}\\
    \bottomrule
    \end{tabular}
    \vspace{-3.6mm}
    }

    \label{prompt_medmcqa}
    \vspace{-4.3mm}
\end{table}

\begin{table}[h]
\centering
    \caption{Prompt for HellaSwag dataset.}
    \scalebox{0.90}{
    \begin{tabular}[c]{@{}p{16.5cm}@{}}
      \toprule
      \textbf{MCQ Prompt:}
     \texttt{The following is the multiple choice question. Please select the correct answer from the options A, B, C, D. For example, if you think the correct answer is A, your response should be 'A'.}\\
      \texttt{Question: How to clean your rv windows and mirrors fast without using any spray. you}\\
      \texttt{A. \textcolor{blue}{also have a bucket that you spray paint a window in.}}\\
      \texttt{B. \textcolor{blue}{can reach for a running water hose and clean the inside of your rv quickly.}}\\
      \texttt{C. \textcolor{blue}{get a wash cloth and you put it under the faucet to get wet and then you rinse it out so it's not soaking.}}\\
      \texttt{D. \textcolor{blue}{meticulously clean the window in the glass shop and then take the plastic off and start taking the hood off.}}\\
    \texttt{Answer:}
    \\
    \\
    \textbf{Open-Style Prompt:} \texttt{Imagine you are provided with a scenario or a partial story taken from everyday life or a common activity. Your task is to continue this story or scenario in a way that makes the most sense based on what typically happens in such situations. Please complete the sentence.}\\
    \texttt{Question: How to clean your rv windows and mirrors fast without using any spray. you}\\
    \texttt{Answer:}\\
    \bottomrule
    \end{tabular}
    \vspace{-3.6mm}
    }

    \label{prompt_hs}
    \vspace{-4.3mm}
\end{table}

\begin{table}[h]
\centering
    \caption{Prompt for OpenbookQA dataset.}
    \scalebox{0.95}{
    \begin{tabular}[c]{@{}p{16.5cm}@{}}
      \toprule
      \textbf{MCQ Prompt:}
     \texttt{The following is the multiple choice question. Please select the correct answer from the options A, B, C, D. For example, if you think the correct answer is A, your response should be 'A'.}\\
      \texttt{Question: what system is needed for a body to get its needed supply of the gas humans breathe in?}\\
      \texttt{A. \textcolor{blue}{the circulatory system}}\\
      \texttt{B. \textcolor{blue}{the digestive system}}\\
      \texttt{C. \textcolor{blue}{the school system}}\\
      \texttt{D. \textcolor{blue}{central nervous system}}\\
    \texttt{Answer:}
    \\
    \\
    \textbf{Open-Style Prompt:} \texttt{Consider common scenarios or outcomes that fit the context of the sentence. Attempt to logically complete the sentences based on common knowledge and reasoning.}\\
    \texttt{Question: what system is needed for a body to get its needed supply of the gas humans breathe in?}\\
    \texttt{Answer:}\\
    \bottomrule
    \end{tabular}
    \vspace{-3.6mm}
    }

    \label{prompt_obqa}
    \vspace{-4.3mm}
\end{table}

\begin{table}[h]
\centering
    \caption{Prompt for PIQA dataset.}
    \scalebox{0.90}{
    \begin{tabular}[c]{@{}p{16.5cm}@{}}
      \toprule
      \textbf{MCQ Prompt:}
     \texttt{The following is the multiple choice question. Please select the correct answer from the options A, B. For example, if you think the correct answer is A, your response should be 'A'.}\\
      \texttt{Question: How do I ready a guinea pig cage for it's new occupants?}\\
      \texttt{A. \textcolor{blue}{Provide the guinea pig with a cage full of a few inches of bedding made of ripped paper strips, you will also need to supply it with a water bottle and a food dish.}}\\
      \texttt{B. \textcolor{blue}{Provide the guinea pig with a cage full of a few inches of bedding made of ripped jeans material, you will also need to supply it with a water bottle and a food dish.}}\\
    \texttt{Answer:}
    \\
    \\
    \textbf{Open-Style Prompt:} \texttt{Consider common scenarios or outcomes that fit the context of the sentence. Attempt to logically complete the sentences based on common knowledge and reasoning.}\\
    \texttt{Question: How do I ready a guinea pig cage for it's new occupants?}\\
    \texttt{Answer:}\\
    \bottomrule
    \end{tabular}
    \vspace{-3.6mm}
    }
    \label{prompt_PIQA}
    \vspace{-4.3mm}
\end{table}

\begin{table}[h]
\centering
    \caption{Prompt for Race dataset.}
    \scalebox{0.90}{
    \begin{tabular}[c]{@{}p{16.5cm}@{}}
      \toprule
      \textbf{MCQ Prompt:}
     \texttt{I will give you a passage with multiple-choice question. Please select the correct answer from the options A, B, C, D. For example, if you think the correct answer is A, your response should be 'A'.}\\
      \texttt{Passage:...}\\
      \texttt{Question: What did Nancy try to do before she fell over?}\\
      \texttt{A. \textcolor{blue}{Measure the depth of the river}}\\
      \texttt{B. \textcolor{blue}{Look for a fallen tree trunk}}\\
      \texttt{C. \textcolor{blue}{Protect her cows from being drowned}}\\
      \texttt{D. \textcolor{blue}{Run away from the flooded farm}}\\
    \texttt{Answer:}
    \\
    \\
    \textbf{Open-Style Prompt:} \texttt{I will give you passage with question. Please, answer the question.}\\
    \texttt{Passage:...}\\
    \texttt{Question: What did Nancy try to do before she fell over?}\\
    \texttt{Answer:}\\
    \bottomrule
    \end{tabular}
    \vspace{-3.6mm}
    }

    \label{prompt_obqa}
    \vspace{-4.3mm}
\end{table}

\begin{table}[h]
\centering
    \caption{Prompt for WinoGrande dataset.}
    \scalebox{0.90}{
    \begin{tabular}[c]{@{}p{16.5cm}@{}}
      \toprule
      \textbf{MCQ Prompt:}
     \texttt{The following is the multiple choice question. Please put the correct words in place of \_. Your response should include only the option without any justification or reasoning. Please select the correct answer from the options A, B. }\\
      \texttt{Question: Sarah was a much better surgeon than Maria so \_ always got the easier cases.}\\
      \texttt{A. \textcolor{blue}{Sarah}}\\
      \texttt{B. \textcolor{blue}{Maria}}\\
    \texttt{Answer:}
    \\
    \\
    \textbf{Open-Style Prompt:} \texttt{Please put the correct words in place of \_. Give only the word that fits the sentence.}\\
    \texttt{Question: Sarah was a much better surgeon than Maria so \_ always got the easier cases.}\\
    \texttt{Answer:}\\
    \bottomrule
    \end{tabular}
    \vspace{-3.6mm}
    }

    \label{prompt_wino}
    \vspace{-4.3mm}
\end{table}

\end{document}